\documentclass[a4paper,preprint,review,authoryear,1p,12pt]{elsarticle}
\usepackage{amssymb}
\usepackage{color}
\usepackage{subfigure}
\usepackage{amsmath}      
\usepackage{array}     
\usepackage[titletoc]{appendix}
\usepackage{amsthm}
\usepackage[figuresright]{rotating}
\usepackage{float}
\usepackage{soul}
\usepackage{algorithmicx}
\usepackage{algpseudocode}
\biboptions{numbers,super,comma,sort&compress}
\usepackage[bookmarksnumbered,colorlinks,linkcolor=red,anchorcolor=blue,citecolor=green]{hyperref}

\usepackage{geometry}
\graphicspath{{figures/}}
\usepackage{multirow} 
\usepackage{booktabs} 
\usepackage{caption}
\usepackage{epstopdf}

\usepackage{graphics}
\usepackage{epsfig}
\usepackage[ruled,linesnumbered]{algorithm2e}

\begin{document}

\begin{frontmatter}
	
	\title{{Dynamic fault detection and diagnosis of industrial alkaline water electrolyzer process with variational Bayesian dictionary learning}}

	\author[label1]{Qi Zhang}
	\author[label1]{Lei Xie  \corref{cor1}}
	\author[label1]{Weihua Xu}
	\author[label1]{Hongye Su}

	\address[label1]
	{State Key Laboratory of Industrial Control Technology, Zhejiang University, 310027 Hangzhou, China}
	\cortext[cor1]{Corresponding author}

	\begin{abstract}

		Alkaline Water Electrolysis (AWE) is one of the simplest green hydrogen production method using renewable energy.  
		AWE system typically yields process variables that are serially correlated and contaminated by measurement uncertainty. 
		A novel robust dynamic variational Bayesian dictionary learning (RDVDL) monitoring approach is proposed to improve the reliability and safety of AWE operation.
		RDVDL employs a sparse Bayesian dictionary learning to preserve the dynamic mechanism information of AWE process which allows the easy interpretation of fault detection results.
		To improve the robustness to measurement uncertainty, a low-rank vector autoregressive (VAR) method is derived to reliably extract the serial correlation from process variables.
		The effectiveness of the proposed approach is demonstrated with an industrial hydrogen production process, and RDVDL can efficiently detect and diagnose critical AWE faults.

	\end{abstract}
	
	\begin{keyword}
		Alkaline water electrolytic, bayesian dictionary learning, process monitoring, data-driven method, Fault detection and diagnosis
	\end{keyword}
	
\end{frontmatter}

\section{Introduction}

	The non-renewable nature of fossil energy is making it progressively depleted and significantly unsustainable\cite{khanReviewRecentOptimization2022}. The utilization of fossil energy is also a key factor causing environmental pollution\cite{wapplerBuildingGreenHydrogen2022}. A large amount of fossil energy consumption increases the concentration of greenhouse gases in the atmosphere, leading to global warming\cite{friedlingsteinGlobalCarbonBudget2022}. With the increasing energy consumption and environmental pollution, it is imperative to expand the scale of renewable energy utilization\cite{parraReviewRoleCost2019}\cite{qureshyEnergyExergyAnalyses2020}.
	Hydrogen, as a renewable energy source, is one of the most capable fuel options for the future, emitting only water vapour as a by-product during the combustion or oxidation process\cite{aminHydrogenProductionRenewable2022}. 
	Hydrogen also has considerable potential as an energy storage medium for storing excess renewable energy such as wind or solar energy\cite{xiaoOptimalOperationWindelectrolytic2020}\cite{abeHydrogenEnergyEconomy2019}.
	Hydrogen energy storage systems are becoming increasingly popular as a relatively inexpensive way to store renewable energy, promising a low-carbon energy system transition\cite{abdinHydrogenEnergyVector2020}. Integration of energy storage systems into renewable energy hubs is gaining attention as a promising hybrid energy system
	\cite{mohammadiOptimalManagementEnergy2018b}. 
	In times of energy surplus, hydrogen is produced and stored using electrolyzers. In times of energy shortage, hydrogen is replenished to the microgrid using fuel cells\cite{hossainModelingPerformanceAnalysis2022}.
	Renewable energy plays an important role in the electricity market, and the integration of hydrogen into the electricity market is making significant progress as one of the pillars of decarbonization efforts in the power system\cite{heselIntegratedModellingEuropean2022}\cite{arsadHydrogenEnergyStorage2022}.Hydrogen participation in the electricity market can build a micro-grid system for multiple energy supply of heat, electricity and hydrogen, effectively enhancing the risk resistance of the new power system\cite{hemmatiThermodynamicModelingCompressed2021}\cite{mirzaeiNovelHybridTwostage2020a}.
	Since hydrogen can be extracted from fuel cells by electrochemical conversion, the hydrogen in fuel cells is used more efficiently than in conventional technologies, producing high quality energy with little waste
	\cite{aminudinOverviewCurrentProgress2023}.
	Due to its high efficiency, high power density, and water-only emissions, proton exchange membrane fuel cells (PEMFC) have been acknowledged as one of the future directions for vehicle power\cite{liRecentAdvancesAnode2023}\cite{wilberforceRecoveryWasteHeat2022}. Vehicles fueled by hydrogen have better fuel refueling frequency and range than current lithium battery-based electric vehicles\cite{kasaeianIntegrationSolidOxide2023}. 
	Fuel cells are well suited for long-distance transportation and produce almost no harmful emissions\cite{weiBibliometricAnalysisSafety2023}. While stability and cost remain major challenges, zero emissions may be the most attractive aspect of the fuel cell vehicle market\cite{pramuanjaroenkijFuelCellElectric2022}. As a result, the use of fuel cells in transportation is seen as promising\cite{fanRecentDevelopmentHydrogen2021}. 

	As low carbon becomes a new goal of industrial development, there has been an increased attention to the hydrogen production\cite{acarSelectionCriteriaRanking2022}. Industrial hydrogen production methods mainly include water electrolysis, methanol steam reforming, and steam catalytic conversion of heavy oil and natural gas \cite{papadisChallengesDecarbonizationEnergy2020}.
	In addition, energy efficiency and environmental cleanliness are advantages of producing hydrogen from biomass feedstock\cite{catumbaSustainabilityChallengesHydrogen2022}. Biophotolysis generates hydrogen by microorganisms with the ability to split water molecules through light energy\cite{ferraren-decagalitanReviewBiohydrogenProduction2021}. In a setting with insufficient nitrogen, microbial fermentation technology converts organic acids produced during anaerobic fermentation into hydrogen\cite{dincerReviewEvaluationHydrogen2015}. The possibility of combining waste treatment with the production of clean fuels has made it a popular research topic\cite{palReviewBiomassBased2022a}.
	Although the cost of hydrogen production by water electrolysis is higher among these methods, water electrolysis is a key technology for decomposing water into hydrogen and oxygen using renewable energy sources (wind and solar power), and therefore water electrolysis is still considered as the future direction of hydrogen production
	\cite{dossantosHydrogenProductionElectrolysis2017}\cite{yangEconomicAnalysisHydrogen2023}. 
	Currently, there are three main technologies for hydrogen production by water electrolysis, alkaline water electrolysis (AWE), proton exchange membrane (PEM) and solid oxide electrolysis (SOEC)\cite{shivakumarOverviewWaterElectrolysis2022}\cite{carmoComprehensiveReviewPEM2013}. 
	
	AWE stands out from other technologies because it is the technology with a more established industry and greater commercial outreach\cite{ursuaHydrogenProductionWater2012}\cite{hoisangKeyCriteriaNextGeneration2022}. 
	The system consists of a pair of electrodes separated by a diaphragm filled with an alkaline solution, usually potassium hydroxide (KOH) at concentrations between 25\% and 30\%\cite{davidAdvancesAlkalineWater2019a}. Water decomposes at the cathode to form $ H_2 $ and releases the hydroxide anion that crosses the diaphragm and combines at the anode to form $ O_2 $\cite{langeTechnicalEvaluationFlexibility2023}.
	The safe operation of AWE system is a challenging task because there are many types of faults that can lead to premature equipment replacement and operational downtime\cite{kheirrouzFaultDetectionDiagnosis2022}. Fault of any component that is not quickly detected and corrected can seriously affect the efficiency of the entire system and cause serious accidents.
	For example, in electrolyzer, membrane electrode assemblies can suffer from different faults: membrane rupture, internal gas leakage, cell overflow or drying, catalyst area poisoning.
	Since these systems are exposed to multiple types of faults and degradation depending on their operating conditions, modeling, degradation studies, fault diagnosis, and fault operation management must be considered. The introduction of hydrogen as a safe and sustainable energy carrier can only be achieved by enhancing electrolyzer technology.
	Detecting and identifying faults in time can considerably improve efficiency and reliability in AWE systems.
	
	To date, water electrolysis modeling has been extensively studied, some technologies based on monitoring pressure, vibration and images have been used to diagnose the hydrogen production process \cite{kheirrouzFaultDetectionDiagnosis2022}. However, little research has been done on fault detection in AWE. 
	A complete AWE hydrogen production plant consists of many components: alkaline electrolyzer, heat exchanger, gas-liquid separator, scrubber, hydrogen-oxygen purifier, electrical energy system, etc. 
	In the process of hydrogen production, status information is collected and stored in real time by different sensors, including temperature sensors, liquid level sensors, pressure sensors, flow meters, gas purity sensors, flow meters, current and voltage sensors, electrolyte concentration sensors.
	Since these data reflect the changes of the process, it becomes a hot topic to study these data by statistical modeling methods. 
	In recent years, data-driven process monitoring methods have been applied to fault detection and diagnosis
	\cite{qinSurveyDatadrivenIndustrial2012a}\cite{wangSurveyTheoreticalResearch2018}.
	Usually, the key idea of data-driven is that the system can access a large amount of data across different processes, and most of these data are highly correlated  . In order to extract the most discriminative features from the raw data, dimensionality reduction methods are highly expected.
	Principal component analysis (PCA), as a representative method of dimensionality reduction technology, has been widely studied for its simplicity and effectiveness in handling large amounts of process data \cite{joeqinStatisticalProcessMonitoring2003a}. 
	
	During the AWE process, the current and voltage data collected by the sensors are easily disturbed by harmonics, and these low-quality data can easily cause misjudgment. In addition, the electrolytic water chemical reaction mechanism and the voltammetric characteristics of alkaline electrolyzer show that the equivalent resistive impedance of alkaline electrolyzer is only related to the tank temperature, and the operating state of electrolyzer shows the correlation of time series.
	The PCA-based approach minimizes the scattering of all projected samples by obtaining an orthogonal projection using the training data\cite{joeqinStatisticalProcessMonitoring2003a}.
	Although the dimensionality of the data is reduced in this process, the structural information between the data is also lost, which may make PCA insensitive to small faults\cite{zhangQualityRelevantProcessMonitoring2022a}.
	Compressed sensing theory proves that signals can be reconstructed from very limited measurements if they are somehow sparsely represented  \cite{gaoDimensionalityReductionCompressive2012}\cite{chenCompressedSensingBased2014}. Therefore, compressed sensing can be regarded as using sparsity to achieve dimensionality reduction. 
	PCA is to solve the orthonormal basis so that the data has the maximum variance after projection, and dictionary learning is to solve the dictionary matrix to make the original matrix sparse enough\cite{tosicDictionaryLearning2011}\cite{kreutz-delgadoDictionaryLearningAlgorithms2003}. 
	Inspired by the principle of sparse dimensionality reduction, we try to build a fault detection model for AWE hydrogen production process from the perspective of Bayesian sparsity to enhance the effectiveness of model monitoring\cite{wrightSparseReconstructionSeparable2009}\cite{paisleyNonparametricFactorAnalysis2009}.

	Operational data from the AWE systems are usually high dimensional with dynamic features, multiple sampling rates, and a mix of continuous and categorical quantities.
	Time series correlation of process data is a popular topic in process monitoring. Dynamic latent variables (DLV) methods represented by dynamic PCA  have been proposed to extract dynamic properties of data\cite{kuDisturbanceDetectionIsolation1995a}. 
	It is interesting to study the dynamics of AWE process, but severe noise perturbations contaminate the data, leading to distorted dynamic relationships\cite{zhangQualityRelevantProcessMonitoring2022a}.
	In a real industrial processes, a priori information on noise variance and sparsity is usually not available and can only be tuned manually. Furthermore, inaccurate parameters can lead to significant performance degradation. As a result, We try to address this problem from a Bayesian nonparametric perspective. Specifically, the beta process, as a conjugate prior to the Bernoulli distribution, ensures that all observations share a sparse subset of factors, where the actual number of dictionary atoms and their relative importance are inferred from the data without a priori knowledge\cite{zhangVariationalBayesianState2023}. The sparse matrix is obtained by variational Bayesian (VB) inference. The advantage of variational inference is the ability to minimize the Kullback-Leibler (KL) Divergence between the approximate posterior distribution and the true posterior distribution while maximizing the sample log-likelihood function. With the help of mean field approximation, the true posterior distribution can be approximated by a factorizable distribution  \cite{bealVariationalBayesianLearning2006}\cite{bleiVariationalInferenceReview2017}.
	
	In this paper, a process monitoring approach for AWE hydrogen production systems based on data-driven method is developed. Specifically, the industrial hydrogen production process is reviewed, ten common faults and collected variables are analyzed. Then several representative dynamic data-driven methods are reviewed. Finally, a Bayesian dictionary learning based dynamic fault detection and diagnosis method is developed for the industrial hydrogen production process. 
	The fault detection method for robust dynamic variational Bayesian dictionary learning (RDVDL) is developed for AWE hydrogen production. The proposed method can properly preserve the structural information of AWE process by sparse dimensionality reduction, and the dictionary matrix is solved by variational  expectation maximization (EM). To improve the robustness of dynamic analysis, a low-rank vector autoregressive (VAR) method is derived to further extract the autocorrelation characteristics of the reconstruction variables. Test results using an industrial hydrogen production process demonstrate the effectiveness of the data-driven approach, and the developed RDVDL can efficiently detect faults and diagnose fault variables.

The remainder of this paper is organized as follows.
Section~\ref{sec:related} reviews the related works.
Section~\ref{sec:algorithms} presents the algorithms. Section~\ref{sec:fault detection} develops process monitoring methods, which are evaluated on Section~\ref{sec:experiments}.  Finally, Section~\ref{sec:conclusions} draws the concluding remarks of our paper.

\section{AWE hydrogen production} \label{sec:related}

	\subsection{AWE hydrogen production system}
	
\begin{figure}[h]
		\centering
		\includegraphics[width=7in]{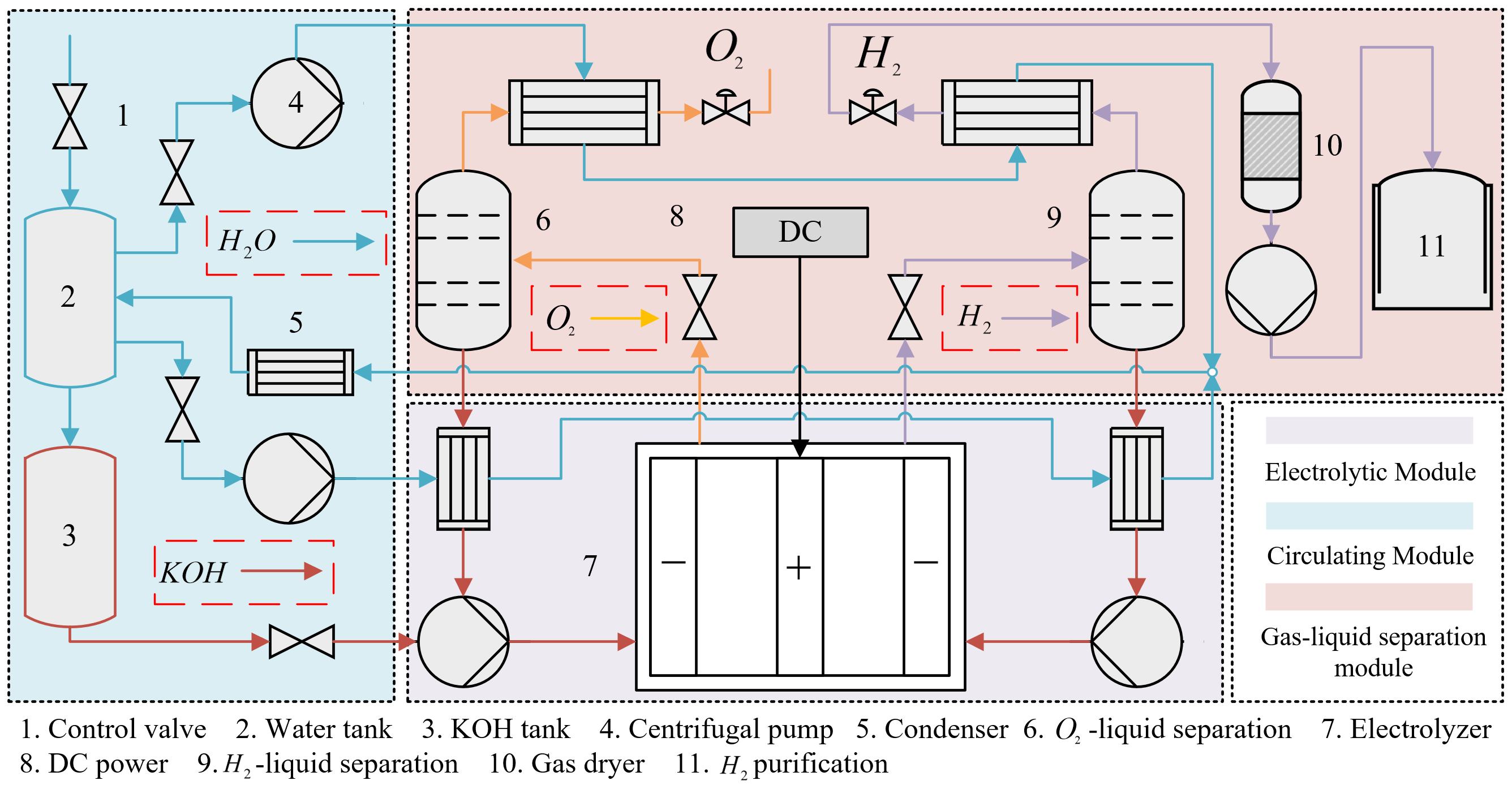}
		\caption{ Flow chart of alkaline electrolysis system.}
		\label{fig.1}
		\end{figure}

	A complete AWE hydrogen production system mainly includes three parts, electrolytic module, gas-liquid separation module, and circulating module. The industrial process of AWE hydrogen production is shown in Figure \ref{fig.1}. 
	The direct current supply powers the alkaline water electrolyzer, and the electrolyte is then decomposed to generate hydrogen and oxygen.
	Hydrogen and oxygen are dissolved in the electrolyte. When the pressure rises to the rated value, the mixtures is sent to the gas-liquid separator through the back pressure valve, where gas and electrolyte are separated.
	 After separation, the gases and liquid go through different routes.
	 The filtered electrolyte is cooled to the working temperature by the condenser, and then the electrolyte is returned to the electrolytic cell through the circulating pump, which can reduce the temperature of the electrolyzer.
	 The separated gas is cooled by the condenser, and the cooled gas enters the gas dryer for dehydration. In order to improve the purity of hydrogen, hydrogen is purified and stored by $ H_2 $ purification equipment.
	 The liquid flows through the circulation pump and filter and then returns to the electrolysis cell from the bottom of the separators. The hydrogen is tested for purity and then collected and stored.
	 During the electrolysis process, the circulation module cools the system while circulating the electrolyte. Pure water is consumed by electrolysis, so it is necessary to supplement water to get electrolyte. The generated electrolyte and the separated electrolyte are circulated into the electrolyzer to maintain the set concentration.

	\subsubsection{Electrolytic module}

	The alkaline electrolyzer module is the most central equipment in the AWE hydrogen production system. As the core equipment of hydrogen production, it directly affects the efficiency of hydrogen production.
	The external part of the alkaline electrolyzer has hydrogen outlet, oxygen outlet and electrolyte inlet. 
	The electrolyte enters the electrolyzer and water is electrolyzed into hydrogen and oxygen in each electrolytic cell. The mixture of hydrogen and electrolyte flows out from the hydrogen side outlet and the mixture of oxygen and electrolyte flows out from the oxygen side outlet.
	
	The internal structure of the electrolyzer is divided into bipolar plate, anode, cathode, iso-osmotic membrane, and sealing ring. The bipolar plate separates the inside of the electrolyzer into several electrolysis chambers.
	The electrode determines the efficiency of hydrogen production, so the electrode material should have the properties of fast hydrogen absorption and dehydrogenation.
	The iso-osmotic membrane prevents gas mixing between the hydrogen side and oxygen side in the electrolysis chamber, and the pressure on the hydrogen side and oxygen side must be equal.
	The electrolyte used in alkaline water electrolyzers is traditionally potassium hydroxide (KOH), with most solutions being 20-30 $wt\%$, due to the optimum conductivity and significant corrosion resistance of stainless steel in this concentration range. 
	Typical operating temperatures and pressures are 70-100${}^\circ C$ and 1-30$bar$, respectively.
	At the cathode water is split to form $ {{\text{H}}_{2}} $ and releasing hydroxide anions which pass through the diaphragm and recombine at the anode to form $ {{\text{O}}_{2}} $ according to the following reactions:
	\begin{align}
		\begin{split}
			& \begin{matrix}
				Cathode:\ \ \ 2{{\text{H}}_{2}}{{\text{O}}_{(\text{l})}}+2{{\text{e}}^{-}}\to {{\text{H}}_{2(\text{g})}}+2O{{H}^{-}}_{(aq)}  \\
			\end{matrix} \\ 
			& \begin{matrix}
				Anode:\ \ \;2O{{H}^{-}}_{(aq)}\to 0.5{{\text{O}}_{2(\text{g})}}+{{\text{H}}_{2}}{{\text{O}}_{(\text{l})}}+2{{\text{e}}^{-}}  \\
			\end{matrix} \\ 
			& \begin{matrix}
				Overallreaction:\ \ \;{{\text{H}}_{2}}{{\text{O}}_{(\text{l})}}\to {{\text{H}}_{2(\text{g})}}+0.5{{\text{O}}_{2(\text{g})}}  \\
			\end{matrix}  
		\end{split}
	\end{align}

\begin{figure}[h]
		\centering
		\includegraphics[width=3in]{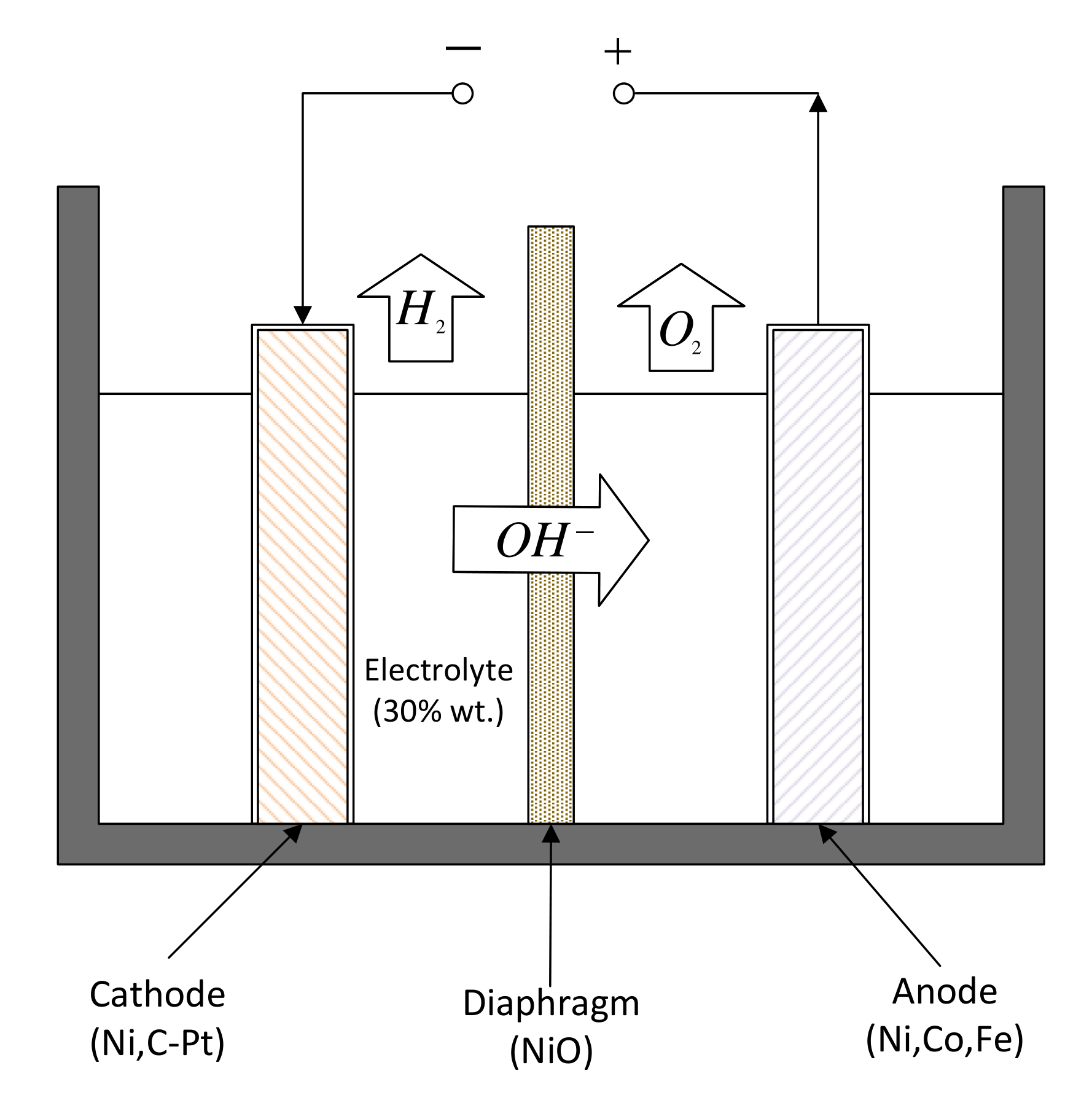}
		\caption{ The chemical reactions in the alkaline electrolyzer.}
		\label{fig.2}
	\end{figure}

	Figure \ref{fig.2} shows the chemical reactions taking place in the electrolyte.
	
\subsubsection{Gas-liquid separation module}
	
	The gas-liquid separation and cooling system is another major component of the AWE hydrogen plant, which mainly consists of three parts: gas-liquid separator, condenser, gas dryer and pressure balancing valve.
	In order to ensure equal pressure operation, the separator is equipped with a differential pressure interface and a level gauge interface for installing the differential pressure and the level gauge, respectively. 
	Real-time monitoring of the liquid level in the gas-liquid separator can be achieved.
	The condenser is used to cool the hydrogen and oxygen after passing through the gas-liquid separator, and the gas dryer is used to further separate the water from the gas. Finally, hydrogen is purified and stored by $ H_2 $ purification equipment.

	The pressure balance between the hydrogen side and the oxygen side is achieved through the pressure balancing valve. When the booster valve is opened, the generated hydrogen and oxygen enter the lower chamber of the balancing valve. When the pressure increases to the control value, the booster valve is closed.
	The balancing valve automatically adjusts the flow rate as the pressure changes, thus achieving the balance of pressure between the upper and lower chambers of the balancing valve.
	
\subsubsection{Circulating module}
	
	The alkaline tank and water tank are external devices independent of the internal circulation part. The water tank serves two purposes. First, the system can be cooled by water circulation. Secondly, the water flows into the KOH tank to configure the electrolyte, which enters the electrolyzer for internal circulation. The alkali tank and water tank are fitted with level meters to monitor the level in the tank.
	
	When the system is operating, the electrolyte begins to circulate to reduce the temperature of the electrolyzer. Likewise, the water also begins to circulate to lower the temperature of the system. The circulation pump is an important equipment to ensure stable electrolyte circulation.
	The product mixtures after separation of hydrogen and oxygen is filtered to remove impurities, and then enters the electrolyzer to start a new cycle.

	\subsection{Summary of different type of faults in AWE electrolyzer.}
	
		Safety has always been a major issue in hydrogen production. The AWE hydrogen production process involves several scientific fields and faults are usually the result of several related factors, which leads to highly correlated parameters that make these systems particularly complex and difficult to control.
	The main faults that may occur in a AWE hydrogen production system are summarized in the Table \ref{Table1}.
	As shown in the table, there are ten common faults in the three parts of the AWE hydrogen production system. In this process, 10 control variables (CV) and 22 process variables (PV) are collected. 
	The control variables are shown in Table \ref{Table2} and the process variables are shown in Table \ref{Table3}.

	\begin{table}[!t]\centering
		\caption{Summary of different type of faults in AWE system.}
		\label{Table1}
		\setlength{\tabcolsep}{5mm}
		\begin{tabular}{c c c} 
			\toprule
			Tag & Fault description & Module \\
			\hline
			1 & Electrolysis cell voltage rise &	Electrolytic module	\\
			2 &	 Electrolyte inter-electrode voltage is abnormal	&	Electrolytic module \\
			3 & Electrolyzer short circuit	&	Electrolytic module  \\
			4 & Reduced gas purity  & Gas-liquid separation module \\ 
			5 & Increased hydrogen oxygen liquid level difference & Gas-liquid separation module\\ 		
			6 & A sharp rise or fall in the hydrogen or oxygen level & Gas-liquid separation module\\ 	
			7 & Unstable pressure & Gas-liquid separation  \\ 	
			8 & Electrolyte stops circulating & Circulating module\\ 	
			9 & Increased electrolyzer temperature & Circulating module \\ 	
			10 & Water tank bubbles & Circulating module  \\
			\bottomrule
		\end{tabular}
	\end{table}

		\begin{table}[!t]\centering
		\caption{Control variable in AWE system.}
		\label{Table2}
		\setlength{\tabcolsep}{1mm}
		\begin{tabular}{c c c} 
			\toprule
			Control variable & Variable description & Unit \\
			\hline
			CV(1) & Hydrogen mass flow  &	$ km^3/h $	\\
			CV(2) &	 Oxygen mass flow	&	$ km^3/h $  \\
			CV(3) & Hydrogen gas-liquid separator liquid level	&	L  \\
			CV(4) & Oxygen gas-liquid separator liquid level  &  L   \\ 
			CV(5) & Water tank return flow & $ km^3/h $  \\ 		
			CV(6) & KOH tank outlet flow & $ km^3/h $  \\ 	
			CV(7) & Electrolytic cell electrode current & A  \\ 	
			CV(8) & Electrolyzer total voltage & V \\ 	
			CV(9) & Electrolyte concentration & wt\%  \\ 	
			CV(10) & Electrolyzer inlet flow & $ km^3/h $  \\
			\bottomrule
		\end{tabular}
	\end{table}
	
			\begin{table}[!t]\centering
		\caption{Process variable in AWE system.}
		\label{Table3}
		\setlength{\tabcolsep}{1mm}
		\begin{tabular}{c c c} 
			\toprule
			Process variable & Variable description & Unit \\
			\hline
			PV(1) & Electrolyzer cell temperature  &	$^{\circ}$C 	\\
			PV(2) &	 Electrolyzer temperature	&	$^{\circ}$C  \\
			PV(3) & Electrolyte level	&	L  \\
			PV(4) & Electrolyzer pressure  & kPa   \\ 
			PV(5) & Hydrogen purity & umol/mol  \\ 		
			PV(6) & Oxygen purity & umol/mol  \\ 	
			PV(7) & Hydrogen separator pressure & kPa  \\ 	
			PV(8) & Oxygen separator pressure & $^{\circ}$C \\ 	
			PV(9) & Hydrogen condenser temperature & $^{\circ}$C \\ 	
			PV(10) & Oxygen condenser temperature & $^{\circ}$C  \\
			PV(11) & Oxygen electrolyte condenser temperature & $^{\circ}$C  \\
			PV(12) & Hydrogen electrolyte condenser temperature & $^{\circ}$C  \\
			PV(13) & Return water condenser temperature & $^{\circ}$C  \\
			PV(14) & KOH tank level & L  \\
			PV(15) & Water tank level & L  \\
			PV(16) & Gas dryer humidity &  \%RH \\
			PV(17) & Gas dryer pressure & kPa \\
			PV(18) & Return pipeline pressure & kPa \\
			PV(19) & Hydrogen separator electrolyte concentration & wt\% \\
			PV(20) & Oxygen separator electrolyte concentration & wt\% \\
			PV(21) & Hydrogen purifier pressure & kPa\\
			PV(22) & Hydrogen purifier purity & umol/mol\\
			\bottomrule
		\end{tabular}
	\end{table}

\subsection{Data-driven process monitoring methods} 

As mentioned above, any single fault in an AWE system may be the result of a combination of factors. Fault diagnosis becomes complex and challenging. 
With the development of sensor technology, various types of high-precision sensing devices are installed in the hydrogen production system to obtain and analyze operating data online, so as to realize the multi-dimensional and real-time collection of the characteristic quantities of the system operating state.

Data-driven process monitoring applies multivariate statistics and machine learning methods to fault detection and diagnosis for industrial process operations and production results. 
The basic idea is to obtain a large amount of data from different operating units of the system, then the characteristics of these data are extracted and correlations are eliminated by dimensional projection, and finally the cause of the fault is detected and diagnosed.

PCA is a basic data-driven process monitoring method, but the PCA method assumes that the data samples are independent in time and does not analyze correlations between variables. In the hydrogen production process, changes in the control variables lead to changes in the corresponding process variables. Due to the closed-loop operation of the circulatory system, the process of hydrogen production is periodic, so it is meaningful to study dynamic data modeling. The dynamic process modeling techniques is briefly reviewed.

\subsubsection{Review of PCA}
For PCA modeling\cite{krestaMultivariateStatisticalMonitoring1991}\cite{joeqinStatisticalProcessMonitoring2003a}, suppose ${{x}_{i}}\in {{\mathbb{R}}^{n\times m}},i=1,2,...,n$ contains $ m $ variables, where each variable has $ n $ observations. The data matrix is constructed as $\mathbf{X}={{\left[ {{\mathbf{x}}_{1}},{{\mathbf{x}}_{2}},\ldots ,{{\mathbf{x}}_{n}} \right]}^{\top}}\in {\mathbb{R}^{n\times m}}$. The covariance matrix $\mathbf{S}\simeq \frac{1}{m-1}{{\mathbf{X}}^{\top}}\mathbf{X}$
is decomposed by eigendecomposition 
\begin{align}
	\mathbf{S}=\mathbf{\hat{S}}\text{+}\mathbf{\tilde{S}}=\mathbf{P}{\mathbf{\Lambda} }{{\mathbf{P}}^{\top}}+\mathbf{\tilde{P}}\mathbf{\tilde{\Lambda} }{{\mathbf{\tilde{P}}}^{\top}}
\end{align}
where $\mathbf{P}\in {\mathbb{R}}^{n\times l}$, $\mathbf{\tilde{P}}\in {{\mathbb{R}}^{n\times \left( n-l \right)}}$ are the principal and residual loadings, respectively. The diagonal matrices $ {\mathbf{{\Lambda}} } $ and $ {\mathbf{\tilde{\Lambda}} } $ contain the principal and residual eigenvalues, respectively.
The matrices $\mathbf{C}=\mathbf{P}{{\mathbf{P}}^{\top}}$ and $\mathbf{\tilde{C}}=\mathbf{\tilde{P}}{{\mathbf{\tilde{P}}}^{\top}}$ are the principal component subspace (PCS) and residual subspace (RS) projection matrices, respectively.
$ \mathbf{T} = \mathbf{X}\mathbf{P} $ is the score vector.
Then two fault detection index are correspondingly constructed.

The $ SPE $ index is defined as the squared norm of the residual vector
\begin{align}
	\text{SPE}\equiv {{\left\| {\mathbf{\tilde{x}}} \right\|}^{2}}={{\mathbf{x}}^{\top}}\mathbf{\tilde{C}}{{\mathbf{\tilde{C}}}^{\top}}\mathbf{x}={{\mathbf{x}}^{\top}}\mathbf{\tilde{C}x}
\end{align}

The $ T_2 $ index in the PCS is defined as
\begin{align}
	{{T}^{2}}={{\mathbf{x}}^{\top}}\mathbf{P}{{\mathbf{\Lambda }}^{-1}}{{\mathbf{P}}^{\top}}\mathbf{x}={{\mathbf{x}}^{\top}}\mathbf{Dx}
\end{align}

Based on the idea of PCA, Partial least squares (PLS), Canonical correlation analysis (CCA) and other methods are proposed for fault detection and diagnosis. Kernel PCA and Kernel PLS have also been used to monitor processes with nonlinear characteristics, and Independent component analysis (ICA) has also been developed to monitor non-Gaussian processes.

\subsubsection{Review of Dynamic PCA}
Dynamic PCA (DPCA) is the most conventional dynamic monitoring method and has been widely used. The main idea is to construct a dynamic matrix to express the time-series correlation of the data\cite{kuDisturbanceDetectionIsolation1995c}\cite{zhangImprovedDynamicKernel2020a}.
Process data are usually collected along time series, so PCA can be extended to time-dependent dynamic processes. The latent variables associated with measurement vectors are extracted by constructing dynamic matrices.
\begin{align}
	\mathbf{x}_{k}^{\top}=\left[ \mathbf{z}_{k}^{\top},\mathbf{z}_{k-1}^{\top}\cdots \mathbf{z}_{k-d}^{\top} \right]
\end{align}
where $ \mathbf{x}_{k} $ is the set of all correlated variables at time k. The scores vector can be calculated
\begin{align}
	{\mathbf{t}_{k}}={{\mathbf{P}}^{\top}}{{\left[ \mathbf{z}_{k}^{\top},\mathbf{z}_{k-1}^{\top}\cdots \mathbf{z}_{k-d}^{\top} \right]}^{\top}}=\underset{i=0}{\overset{d}{\mathop{\sum }}}\,\mathbf{P}_{i}^{\top}{\mathbf{z}_{k-i}}\equiv \mathbf{A}({{q}^{-1}}){\mathbf{z}_{k}}
\end{align}
where ${{\mathbf{P}}^{\top}}= \left[\mathbf{P}_{0}^{\top} , \mathbf{P}_{1}^{\top} \cdots \mathbf{P}_{d}^{\top}\right]$, $\mathbf{A}({{q}^{-1}})=\sum _{i=0}^{d}\mathbf{P}_{i}^{\top}{{q}^{-i}}$ is the matrix polynomial, $ {{q}^{-1}} $ is the backward shift operator.

\subsubsection{Review of Dynamic-inner PCA}
The purpose of Dynamic-inner PCA (DiPCA) is to obtain the inner dynamics of the data to obtain the predicted relationship between past and present variables\cite{dongDynamicLatentVariable2018}.
\begin{align}
	\mathbf{t}_k={{\beta }_{1}}\mathbf{t}_{k-1}+\cdots +{{\beta }_{s}}\mathbf{t}_{k-1}+{\mathbf{r}_{k}}
\end{align}
where $ \mathbf{t}_k= {\mathbf{x}}_k^{\top}{\mathbf{P}}$, $ \beta_i $ is regression coefficient. The regression model can be derived as
\begin{align}
	{\mathbf{\hat{t}}_{k}}={{\mathbf{x}}_{k-1}}\mathbf{p}{{\beta }_{1}}+\cdots +{{\mathbf{x}}_{k-s}}\mathbf{p}{{\beta }_{s}}=\left[ {{\mathbf{x}}_{k-1}},\ldots ,{{\mathbf{x}}_{k-s}} \right]\left( \beta \otimes \mathbf{p} \right)
\end{align}
where $\beta \otimes \mathbf{p}$ is the Kronecker product. The objective function of DiPCA can be obtained
\begin{align}
	\begin{matrix}
		{} & \underset{\mathbf{p},\beta }{\mathop{\max }}\,\ \ \mathbf{t}_{k}^{\top }{{{\mathbf{\hat{t}}}}_{k}}=\underset{j=1}{\mathop{\overset{k}{\mathop{\sum }}\,}}\,{{\beta }_{i}}{{\mathbf{p}}^{\top }}\mathbf{X}_{k}^{\top }{{\mathbf{X}}_{k-i}}\mathbf{p}  \\
		{} & \text{s}\text{.t}\text{.}\ \parallel \mathbf{p}\parallel =1\ \text{and}\ \parallel \beta \parallel =1  \\
	\end{matrix}
\end{align}

\subsubsection{Review of Dynamic-inner CCA}
Dynamic-inner CCA (DiCCA)\cite{dongNewDynamicPredictive2020} is developed to extract dynamic latent variables $ {\mathbf{t}_{k}}={\mathbf{x}}_k^{\top}{\mathbf{P}} $. The predictor variables can be expressed as
\begin{align}
	{{\mathbf{\hat{t}}}_{k}}=\underset{i=1}{\overset{k}{\mathop \sum }}\,{{\beta }_{i}}{{\mathbf{t}}_{k-i}}=\underset{i=1}{\overset{k}{\mathop \sum }}\,{{\beta }_{i}}{{\mathbf{X}}_{k-i}}\mathbf{p}
\end{align}

 In order to obtain the latent variables of the relevant dynamics, DiCCA looks for $ {\mathbf{t}_{k}} $ so that it is most relevant to predict variable $ {{\mathbf{\hat{t}}}_{k}} $.

The objective function of DiPCA can be obtained
\begin{align}
	\begin{matrix}
		& \underset{\mathbf{p},\beta }{\mathop{\min }}\,\ \ J=\parallel {{\mathbf{t}}_{k}}-{{{\mathbf{\hat{t}}}}_{k}}{{\parallel }^{2}} \\ 
		& \text{s}\text{.t}\text{.}\ \mathbf{t}_{k}^{\top }{{\mathbf{t}}_{k}}=1 \\ 
	\end{matrix}
\end{align}

\section{Dynamic Variational Inference Dictionary Learning} \label{sec:algorithms}

\subsection{Bayesian Dictionary Learning}

The current dynamic monitoring models are mainly based on linear mapping analysis and thus lack probabilistic interpretability. In fact, the data acquisition process is susceptible to noise interference, which can cause serious errors for linear mapping. Probabilistic models represent process variation in terms of data distributions and are therefore robust to missing values. In addition, the sparse model can properly eliminate the effects of noise to improve the reliability of the model. 
As a result, this work builds a dynamic process monitoring model from Bayesian dictionary learning.

Collect $ P $ variables at time length $ N $ to form a data matrix  $\boldsymbol{X}\in {{\mathbb{R}}^{P\times N}}$, where each $ {\boldsymbol{x}_{i}} \in {{\mathbb{R}}^{P\times 1}} $, the Bayesian dictionary matrix is $\boldsymbol{D}\in {{\mathbb{R}}^{P\times K}}$, the vectors of coefficients ${\boldsymbol{w}_{i}}\in {{\mathbb{R}}^{K\times1}}$. The dictionary learning is established as
\begin{align}
	{\boldsymbol{x}_{i}}=\boldsymbol{D}{\boldsymbol{w}_{i}}+{\boldsymbol{\epsilon }_{i}}
\end{align}
where ${\boldsymbol{\epsilon}_{i}}\in {{\mathbb{R}}^{P\times1}}$ is a residual term that encompasses noise and deviation from the linear factor model.

The sparsity can be obtained by the vectors $ {\boldsymbol{w}_{i}} $, which is enforced by placing a beta-Bernoulli prior on $ {\boldsymbol{w}_{i}} $. The approximate sparsity can be generated as
\begin{align}
	\begin{split}
		&{\boldsymbol{z}_{i}}\sim \underset{k=1}{\overset{K}{\mathop \prod }}\,\text{Bernoulli}({{\pi }_{k}})  \\
		&\boldsymbol{\pi} \sim \prod _{k=1}^{K}\text{Beta}({{a}_{0}}/K,{{b}_{0}}(K-1)/K)
	\end{split}
\end{align}
where ${{\pi }_{k}}$ is the $ k $-th component of $ \boldsymbol{\pi } $.
Coefficient $ {\boldsymbol{w}_{i}} $ is expressed as the Hadamard product of two vectors $ {\boldsymbol{z}_{i}} $ and $ {\boldsymbol{s}_{i}} $, where $ {\boldsymbol{z}_{i}} $ is imposed as a binary sparse vector and $ {\boldsymbol{s}_{i}} $ is the full-information representation. The Bayesian dictionary learning can be further expressed as:
\begin{align}
	{{\boldsymbol{x}}_{i}}=\boldsymbol{D}\left( {\boldsymbol{z}_{i}}\odot {\boldsymbol{s}_{i}} \right)+{\boldsymbol{\epsilon }_{i}}
\end{align}
where $\odot $ represents the Hadamard vector product.

The role of sparse binary vectors $ {\boldsymbol{z}_{i}} $ is to select a subset of the columns of $ \boldsymbol{D} $ for representing $ {\boldsymbol{x}_{i}} $, which results in the following generative process for observation
$ i = 1, . . . , N $
\begin{align}
	\begin{split} 
		&{\boldsymbol{d}_{k}}\sim\mathcal{N}(0,{{P}^{-1}}{{\boldsymbol{I}}_{P}}) \\ 
		&{\boldsymbol{s}_{i}}\sim N(0,\gamma _{s}^{-1}{{\boldsymbol{I}}_{K}})   \\
		&{\boldsymbol{\epsilon }_{i}}\sim\mathcal{N}(0,\gamma _{\epsilon }^{-1}{{\boldsymbol{I}}_{P}}) \\ 
		&{{\gamma }_{s}}\sim\text{Gamma}({{c}_{0}},{{d}_{0}}) \\ &{{\gamma }_{\epsilon }}\sim\text{Gamma}({{e}_{0}},{{f}_{0}})\\
		&{{\pi }_{k}}\sim\text{Beta}(\frac{{{a}_{0}}}{K},\frac{{{b}_{0}}(K-1)}{K})  \\ 
	\end{split}
\end{align}
where ${\boldsymbol{d}_{k}}$ represents the $ k $-th column of $ \boldsymbol{D} $, $ {\bf I}_{P}$ and $ {\bf I}_{K} $ represent the identity matrix of size $ P\times P $ and $ K\times K $ respectively.
The constants $ \{a_{0}, b_{0}, c_{0}, d_{0}, e_{0}, f_{0}\}=\Gamma $ denote hyperparameters, which we collect in vector $ \Gamma $. 

\subsection{Variational Bayesian Inference}
In this section, we derive a VB algorithm to perform fast inference for the Bayesian dictionary learning. 
All latent variables denote as $ \boldsymbol{\theta} =\left\{ \mathbf{D},\boldsymbol{s},\boldsymbol{z},\boldsymbol{\pi} ,{{\gamma }_{s}},{{\gamma }_{\epsilon }} \right\} $, the objective is to find the posterior distribution $p(\boldsymbol{\theta} |\boldsymbol{y})$. According to the mean field theory, approximate $p(\boldsymbol{\theta} |\boldsymbol{y})$ by $q(\mathbf{D},\boldsymbol{s},\boldsymbol{z},\boldsymbol{\pi} ,{{\gamma }_{s}},{{\gamma }_{\epsilon }})$ which has a factorized form over the hidden variables $ \boldsymbol{\theta} $
\begin{align}
	q(\mathbf{D},\boldsymbol{s},\boldsymbol{z},\boldsymbol{\pi} ,{{\gamma }_{s}},{{\gamma }_{\epsilon }})=q(\mathbf{D})q(\boldsymbol{z})q(\boldsymbol{s})q(\boldsymbol{\pi} )q({{\gamma }_{s}})q({{\gamma }_{\epsilon }})
\end{align}

The lower bound is correspondingly written as
\begin{align}
	\begin{split}
		&\mathcal{L}={{{E}}_{q}}[\log p(\boldsymbol{x},\mathbf{D},\boldsymbol{s},\boldsymbol{z},\boldsymbol{\pi} ,{{\gamma }_{s}},{{\gamma }_{\epsilon }})]\\
		&-{{{E}}_{q}}[\log q(\mathbf{D},\boldsymbol{s},\boldsymbol{z},\boldsymbol{\pi} ,{{\gamma }_{s}},{{\gamma }_{\epsilon }})]
	\end{split}
\end{align}
Maximizing the lower bound is equivalent to minimizing the KL divergence.

The optimal distribution over the parameters that maximizes the lower bound, the expectation is taken under the distribution of $ {q_{\boldsymbol{\theta }}} $. The approximate posterior distributions have the following forms:
\begin{align}
	\begin{split}
		& q({{d}_{k}})=\mathcal{N}({\boldsymbol{\mu }_{k}},{{\boldsymbol{\Sigma}}_{k}}),q({{s}_{ik}})=\mathcal{N}({{\nu }_{ik}},{{\text{ }\!\!\Omega\!\!\text{ }}_{ik}}) \\ 
		& q({{z}_{ik}})=\text{Bernoulli}({{\eta }_{ik}}),q({{\pi }_{k}})=\text{Beta}({{\tau }_{1k}},{{\tau }_{2k}}) \\ 
		& q({{\gamma}_{s}})=\text{Gamma}({c}',{d}'),q({{\gamma }_{\epsilon }})=\text{Gamma}({e}',{f}') 
	\end{split}
\end{align}
where $ s_{ik}, \nu_{ik}, \Omega_{ik}, z_{ik} $ are the $ k $-th dictionary atom in the $ i $-th patch.

The solution of the dictionary is through the variational Bayes method, which is divided into variational E-step and variational M-step. The specific solution steps are shown in the appendix.

\section{Dynamic Monitoring Modeling of Hydrogen Production Process}  \label{sec:fault detection}

\subsection{Offline Modeling}
In the offline Bayesian dictionary learning stage, the constructed dictionary is used to design the process monitoring scheme. Specially, the reconstruction error resulting from the training data is calculated to determine the control limits. In this work, the kernel density estimation (KDE) method is used to calculate the control limit, which is a non-parametric method to build an unknown density function.
\begin{align} \label{17}
	{{{\mathop{\hat{f}}}\,}_{h}}\left( e \right)=\frac{1}{N}\underset{i=1}{\overset{N}{\mathop{\sum }}}\,{{K}_{h}}\left( e-{{e}_{i}} \right)=\frac{1}{Nh}\underset{i=1}{\overset{N}{\mathop{\sum }}}\,{{K}_{h}}\left( \frac{e-{{e}_{i}}}{h} \right)
\end{align}
where $ {{K}_{h}} $ is a non-negative kernel function and $ h $ is a relative smoothing parameter called bandwidth.  $ \mathbf{e} $ is the reconstruction error, to construct the density estimate accordingly. The confidence limit $ CL $ can be obtained with a confidence level $ \alpha = 0.95 $.

After the dictionary $ \mathbf{D} $ is obtained through the VB inference update, the newly collected samples are reconstructed through the obtained dictionary.
Specifically, when a new test sample $ \mathbf{y}_{i} $ is collected, the sparse encoding $ \hat{\boldsymbol{w}}_{i} $ can be obtained
\begin{align}
	\begin{split} \label{18}
		& {{{\hat{\boldsymbol{w}}}}_{i}}=\underset{\boldsymbol{w}}{\mathop{\arg \min }}\,\parallel {\mathbf{y}_{i}}-\mathbf{D}{\boldsymbol{w}_{i}}\parallel _{2}^{2} \\ 
		& s.t.\parallel {\boldsymbol{w}_{i}}{{\parallel }_{1}}\le T \\ 
	\end{split}
\end{align}
where the sparse code $ {{{\hat{\boldsymbol{w}}}}_{i}} $ can be obtained by OMP algorithm. The reconstructed samples matrix $ \boldsymbol{\Psi} $ can be obtained by sparse coding $ {{{\hat{\boldsymbol{w}}}}_{i}} $

\begin{align}  \label{20}
	\boldsymbol{\Psi} =\sum\limits_{i}^{n}{\mathbf{D}{{{\hat{\boldsymbol{w}}}}_{i}}}
\end{align}

\subsection{Reconstructed Samples Dynamic Modeling}
As the dynamic relationship of the data in the reconstructed sample matrix $ \boldsymbol{\Psi} $ is not explained, it is necessary to build a dynamic model to characterize the autocorrelation within the reconstructed matrix $ \boldsymbol{\Psi} $. For the reconstructed sample $ \boldsymbol{\psi} $ at time $ t $ with a time interval $ T $, it is reasonable to describe $ \boldsymbol{\psi} $ with a vector autoregressive (VAR) model:
\begin{align} \label{21}
	{\boldsymbol{\psi}_{t}}=\underset{k=1}{\overset{d}{\mathop \sum }}\,{\mathbf{A}_{k}}{\boldsymbol{\psi} _{t-k}}+{{\boldsymbol{\epsilon }}_{t}}, \qquad t=d+1,...,T
\end{align}
where ${\mathbf{A}_{k}}\in {{\mathbb{R}}^{n_{\boldsymbol{\psi}}\times n_{\boldsymbol{\psi}}}},k=1,2,...,d$ represents the coefficient matrix of the vector autoregressive model, ${\boldsymbol{\epsilon }_{t}}$ is the Gaussian noise, $ d $ is the time lags.
For convenience, $ \boldsymbol{\psi} $ can further expressed as
\begin{align}\label{5}
	{\boldsymbol{\psi}_{t}}\approx \underset{k=1}{\overset{d}{\mathop \sum }}\,{{\mathbf{A}}_{k}}{\boldsymbol{\psi}_{t-k}}={{\mathbf{A}}^{\top}}{{\mathbf{V}}_{t}},\text{ }\!\!~\!\!\text{ }t=d+1,...,T,
\end{align}
where $\mathbf{A}={{\left[ {\mathbf{A}_{1}^\top },...,{\mathbf{A}_{d}^\top} \right]}}\in {{\mathbb{R}}^{(n_{\boldsymbol{\psi}} \times d)\times n_{\boldsymbol{\psi}}} } $, ${{\mathbf{V}}_{t}}=\left[ \boldsymbol{\psi}_{t-1}^{\top},...,\boldsymbol{\psi}_{t-d}^{\top} \right] \in {{\mathbb{R}}^{(n_{\boldsymbol{\psi}} \times d)}}$. 

Studying low-rank VAR can appropriately improve the robustness of the model, L1 norm regularization is imposed to learn the low-rank projection in the regression model
\begin{align}\label{6}
	\mathbf{\hat{A}}=\underset{\mathbf{A}}{\mathop{\min }}\,\frac{1}{2}\sum\limits_{k=1}^{t}{\|\mathbf{Z}-\mathbf{QA}\|_{2}^{2}}+\lambda {{\left\| \mathbf{A} \right\|}_{1}}
\end{align}
where $ \mathbf{Z} $ and $ \mathbf{Q} $ are defined as $\mathbf{Z}=\left[ \mathbf{\boldsymbol{\psi}}_{d+1}^{\top},...,\mathbf{\boldsymbol{\psi}}_{T}^{\top} \right]\in {{\mathbb{R}}^{\left( T-d \right)\times n_{\boldsymbol{\psi}}}}$, $ \mathbf{Q} ={{\left[ {{\mathbf{v}}_{d+1}^{\top}},...,{{\mathbf{v}}_{T}^{\top}} \right]}}\in {{\mathbb{R}}^{(T-d)\times (n_{\boldsymbol{\psi}} \times d)}}$, respectively.

The rank of $ \mathbf{A} $ is explicitly decided by constraining the rank of $ \mathbf{A} $ to be $s<\min \left( n,k \right)$ 
\begin{align} \label{7}
	\begin{split}
		& \mathbf{\hat{A}}=\underset{\mathbf{X}}{\mathop{\min }}\,\|\mathbf{Z}-\mathbf{QA}\|_{F}^{2} \\  
		& s.t.\text{  }rank\left( \mathbf{A} \right)\le s 
	\end{split}
\end{align}

\subsection{Online Monitoring}

	After extracting the dynamic coefficient matrix from the reconstructed matrix $ \boldsymbol{\psi} $, it is of interest to establish the fault detection statistic separately from the dynamic and static relationships of $ \boldsymbol{\psi} $. 
 	Specifically, modeling static and dynamic relationships separately helps to reveal the autocorrelation of data, which in turn improves fault detection performance.

With the extracted dynamic coefficient matrix , the $ T_{d}^{2} $ statistic is constructed as follow
\begin{align}
	\begin{split}
		T_{d}^{2}={{\left\| \left( \mathbf{I}-\mathbf{\hat{A}D} \right)\mathbf{y}  \right\|}_{2}^{2}}
	\end{split}
\end{align}
where $ \mathbf{y} $ is a new online sample.

Then, static statistics are built from the residuals of the new online sample $ \mathbf{y} $ and the reconstructed sample $ \boldsymbol{\psi} $.
\begin{align}
	T_{s}^{2}=\left( \mathbf{y}-\boldsymbol{\psi}  \right)\boldsymbol{\Lambda} {{\left( \mathbf{y}-\boldsymbol{\psi}  \right)}^{\top }}
\end{align}
where $ \boldsymbol{\Lambda} = \mathbf{Q}^{\top }\mathbf{Q} $.

\subsection{Fault diagnosis}

When a fault is detected in the system, it is important to diagnose the variable that led to the fault. When a fault signal is collected by one of the sensors ${y_{i}}$ in the AWE hydrogen production process, reconstructing the fault variable along the direction of the variable can be expressed as

\begin{align}
	{\mathbf{y}_i}= {\boldsymbol{\psi}_{i}}-{{\xi }_{i}}{f}
\end{align}
where $ {\boldsymbol {\psi}_{i}} $ is the fault-free part of the measurement, $ {{\xi }_{i}} $ is the direction of the fault, $ {{\xi }_{i}}{f} $ is the faulty part, $ {f} $ is the magnitude of the fault. The fault detection index for reconstructing variables through the dictionary matrix  can be written as

\begin{align} \label{20}
	\text{Index}\left( {{\mathbf{y}}_{i}} \right)=\mathbf{y}_{i}^{\top}\boldsymbol{\Phi} {{\mathbf{y}}_{i}}=\|{{\mathbf{y}}_{i}}\|_{\boldsymbol{\Phi} }^{2}=\|\boldsymbol{\psi}-{{\xi }_{i}}{{f}_{i}}\|_{\boldsymbol{\Phi}}^{2}
\end{align}

The purpose of refactoring is to find the value that minimizes $ {{f}_{i}} $. Taking the derivation of \eqref{20}, 
${{f}_{i}}={{\left( \xi _{i}^{\top}{\boldsymbol{\Phi}}{{\xi }_{i}} \right)}^{-1}}\xi _{i}^{\top}{\boldsymbol{\Phi}}{\boldsymbol{\psi}}$ can be obtained. 
The reconstruction-based contribution $ \text{RBC}_{i}^{\text{Index}} $ along the fault direction can be expressed as

\begin{align}
	\text{RBC}_{i}^{\text{Index}}=\|{{\xi }_{i}}{{f}_{i}}\|_{\boldsymbol{\Phi}}^{2}
	={\boldsymbol{\psi}}^{\top}{\boldsymbol{\Phi}}{{\xi }_{i}}{{\left( \xi _{i}^{\top}{\boldsymbol{\Phi}}{{\xi }_{i}} \right)}^{-1}}\xi _{i}^{\top}{\boldsymbol{\Phi}}\boldsymbol{\psi}
\end{align}

In this work, for the $ i_{th} $ variable, for $ k $ samples, the contribution of the fault can be expressed as

\begin{align}
	\label{f1}
	\text{RB}{{\text{C}}_{i,k}}=\frac{{{\left( \xi _{i}^{\top }{\boldsymbol{\Phi}}{{\mathbf{y}}_{i}} \right)}^{2}}}{\xi _{i}^{\top}{\boldsymbol{\Phi}}{{\xi }_{i}}}
\end{align}
where $ {{\xi }_{i}} $ is the $ j_{th} $ column of the identity matrix $ \mathbf{I}_m $, $ {\boldsymbol{\Phi}} = \left( \mathbf{I}-\mathbf{\hat{A}D} \right) $.

In summary, the implementation steps of the RDVDL algorithm is summarized as Algorithm \ref{algorithm1}.

\begin{algorithm} [H]
	\caption{RDVDL for process monitoring}\label{algorithm1}
	\SetKwInOut{Input}{input}\SetKwInOut{Output}{output}
	\Input{ Training data $\mathbf{X}$ and test data $\mathbf{Y}$}
	\Output{Fault detection results}
	Initialize hyperparameters $ \Gamma $, data standardization \;
	{	\While{Stopping criterion is not met}
		{\For {$ k = 1 $ to $ K $}
			{Update $ q(\boldsymbol{d}_{k}) $ with $ \boldsymbol{\Sigma}_{k} $ and $ {\boldsymbol{\mu }_{k}} $ using Eq. \eqref{12}  \;
				\For{$ i=1,...,N $}
				{Update $ q(s_{ik}) $ with $ {{\Omega}_{ik}} $ and $ {{\nu }_{ik}} $ using Eq. \eqref{13}  \;
					Update $ q(\boldsymbol{\pi}) $ with $ {{\tau }_{1k}} $ and $ {{\tau }_{2k}} $ using \eqref{14}  \;
					Update $ {{z}_{ik}} $ using Eq. \eqref{9};
				}
			}
			Update  $ {c}', {d}',{e}',{f}' $ using Eq. \eqref{15} \;}
	}
	Sparse encoding $ \hat{\boldsymbol{w}}_{i} $ is calculate by $ \boldsymbol{y}_{i} $ using Eq. \eqref{18} \;
	Reconstructed matrix $ \boldsymbol{\Psi} $ is calculate using Eq. \eqref{20} \;
	$ \mathbf{\hat{A}} $ is calculate to bulid the statistic $ T_{d}^{2} $ and $ T_{s}^{2} $\;
	Confidence limit $ CL $ is calculate using Eq. \eqref{17} to to detect faults \;
	The contribution of the fault is calculate using Eq. \eqref{f1}.
\end{algorithm}

\section{Case studies} \label{sec:experiments}

The specific hydrogen production process is described in Section \ref{sec:related}. Since the sensor acquisition system realizes the real-time perception of the system operating status and characteristics, the data-driven process monitoring technology has been applied.
In this section, three faults in the three modules of the hydrogen production process are analyzed.
All three faults are introduced from the 201st sample, based on these three faults, fault detection effects of several data-driven methods mentioned above in the hydrogen production process are experimentally analyzed. Meanwhile, the RDVDL method proposed in this work is validated in fault detection and diagnosis.

\subsection{Electrolysis cell voltage rise}
Fault 1 is the voltage rise in the electrolysis. This fault usually occurs due to the blockage of the electrolysis cell, which hinders the flow of ions in the cell. Since the underlying cause of this fault is electrolyte impurities or reduced electrolyte concentration, the affected modules involve circulation modules and electrolysis modules, so multiple sensors are affected. 

\begin{figure}[H]
	\centering
	\subfigure[DPCA.]{
		\includegraphics[width=0.45\textwidth,height=0.3\textwidth]{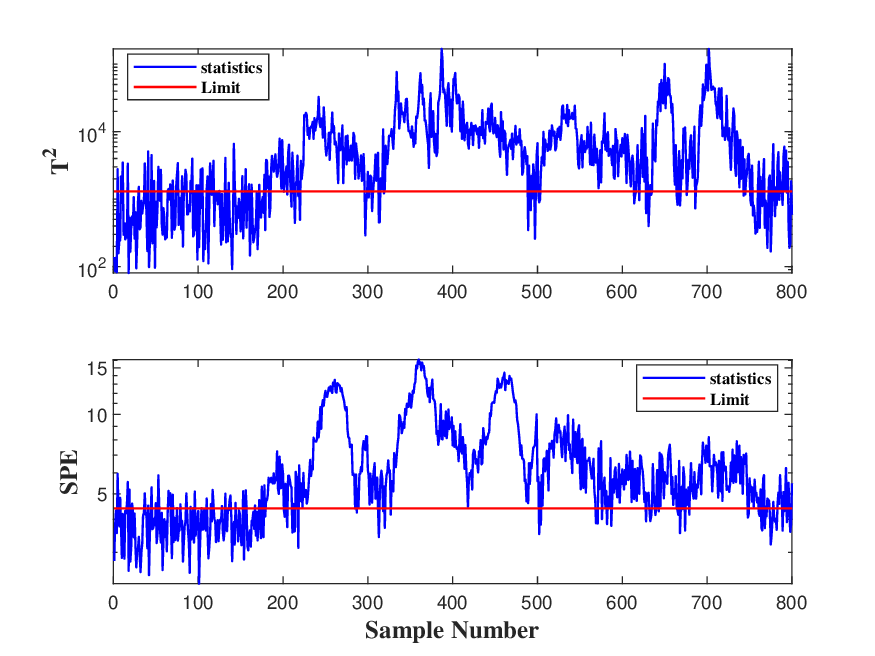}
	}
	\quad
	\subfigure[DiPCA.]{
		\includegraphics[width=0.45\textwidth,height=0.3\textwidth]{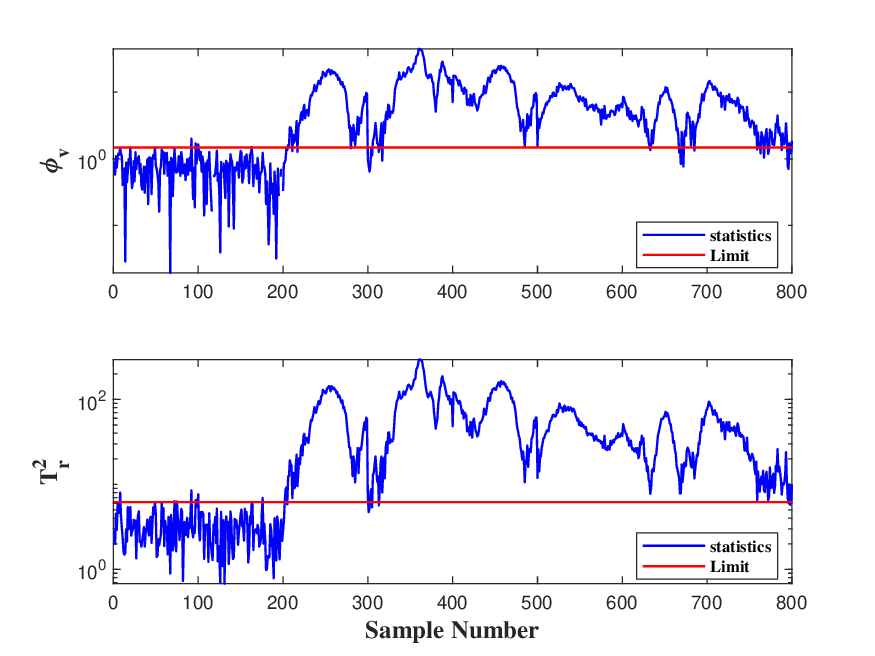}
	}
	\vspace{-1\baselineskip}
	\subfigure[DiCCA.]{
		\includegraphics[width=0.45\textwidth,height=0.3\textwidth]{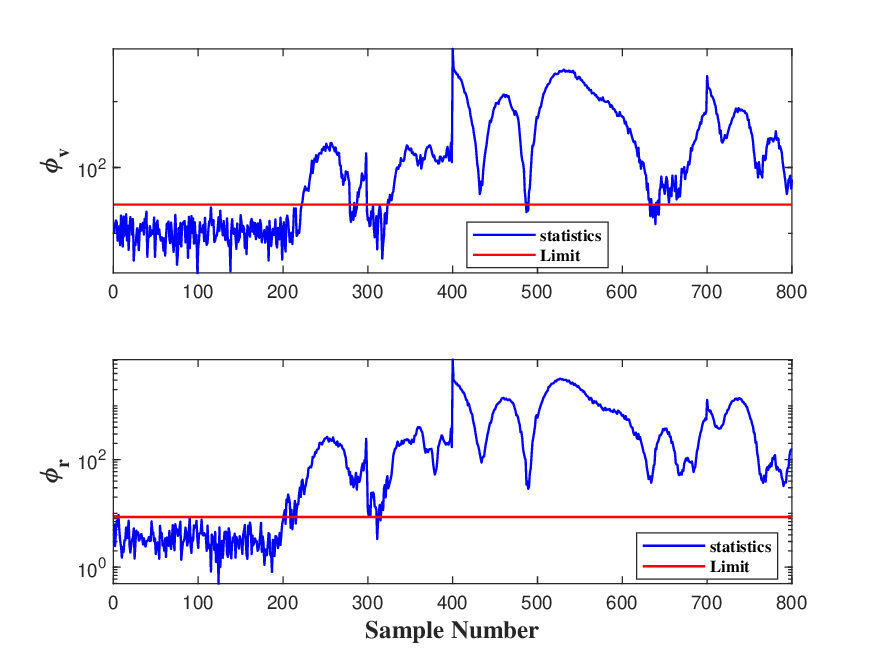}
	}
	\quad
	\subfigure[RDVDL.]{
		\includegraphics[width=0.45\textwidth,height=0.3\textwidth]{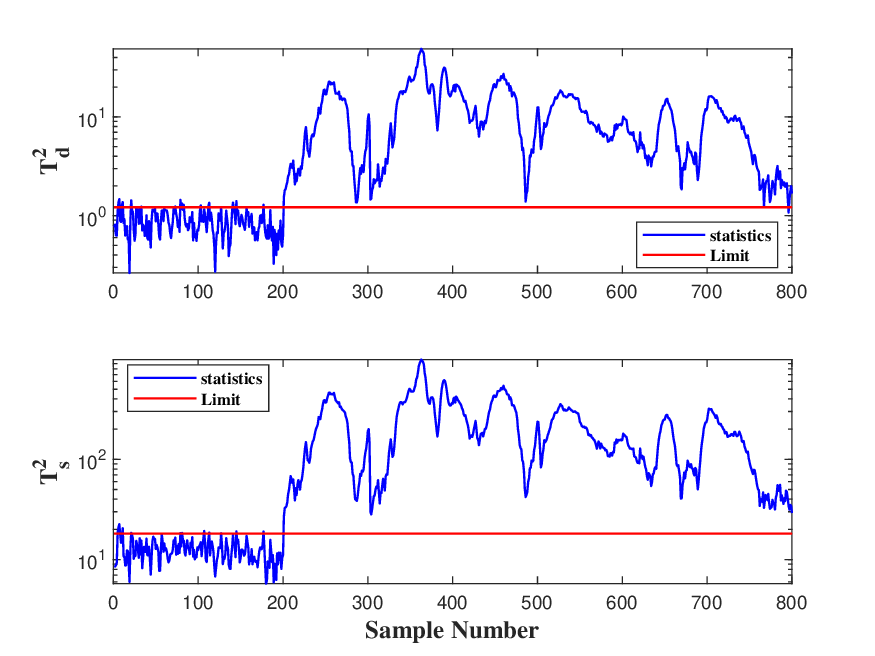}
	}
	\caption{Fault detection result of fault 1.}
	\label{fig.3}
\end{figure}

The results of the fault detection are shown in Figure \ref{fig.3}. The blue line indicates the fault statistic, which presents the process variation with the fault; the red line is the confidence limit, the statistic above the confidence limit indicates that the fault is detected.
The test results show that all four methods can detect faults, indicating that a data-driven process monitoring methods can effectively detect faults in the hydrogen production process.

However, due to the influence of noise, normal samples are easily misidentified as faults. The three compared methods are affected by noise to varying degrees, and there is a high false alarm rate. Since RDVDL takes into account the autocorrelation of online samples, dynamic relationships between data are extracted for building statistics. Therefore, RDVDL has better fault detection effect.

\begin{figure}[H]
	\centering
	\includegraphics[width=0.85\textwidth,height=0.5\textwidth]{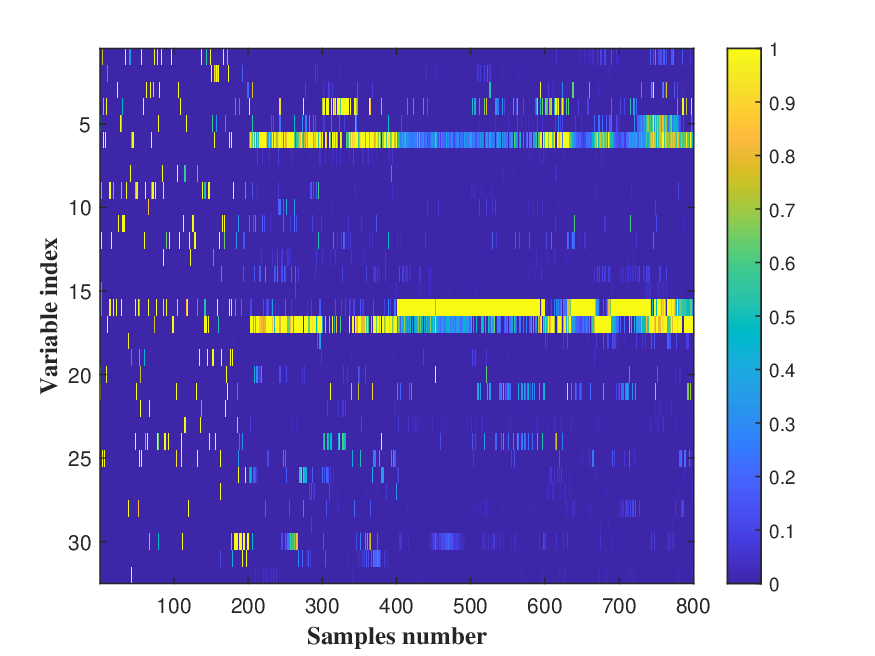}
	\caption{Fault diagnosis result of fault 1.}
	\label{fig.4}
\end{figure}

When a fault is detected, it is necessary to identify the fault variables that are abnormal. Fault 1 may cause abnormal changes in several variables, but the main change should be a change in electrolyte quality. The fault contribution map built using RDVDL is shown in Figure \ref{fig.4}. The diagnosis results show that the variables affected by the fault are variables 6, 16 and 17.

\subsection{Increased hydrogen oxygen liquid level difference}

The increase of the hydrogen-oxygen liquid level difference will cause the pressure to be unbalanced, and the lye in the separator will carry the gas back to the lye circulating pump, which will cause damage to the circulating pump and affect the safe operation of the electrolyzer.
This fault can be caused by several factors. In the actual maintenance process, the pipeline of the hydrogen-oxygen separator is blocked, the balance valve is damaged, the electrolyte concentration is too high, and the leakage of the check valve of the circulation system will cause this fault.

\begin{figure}[H]
	\centering
	\subfigure[DPCA.]{
		\includegraphics[width=0.45\textwidth,height=0.3\textwidth]{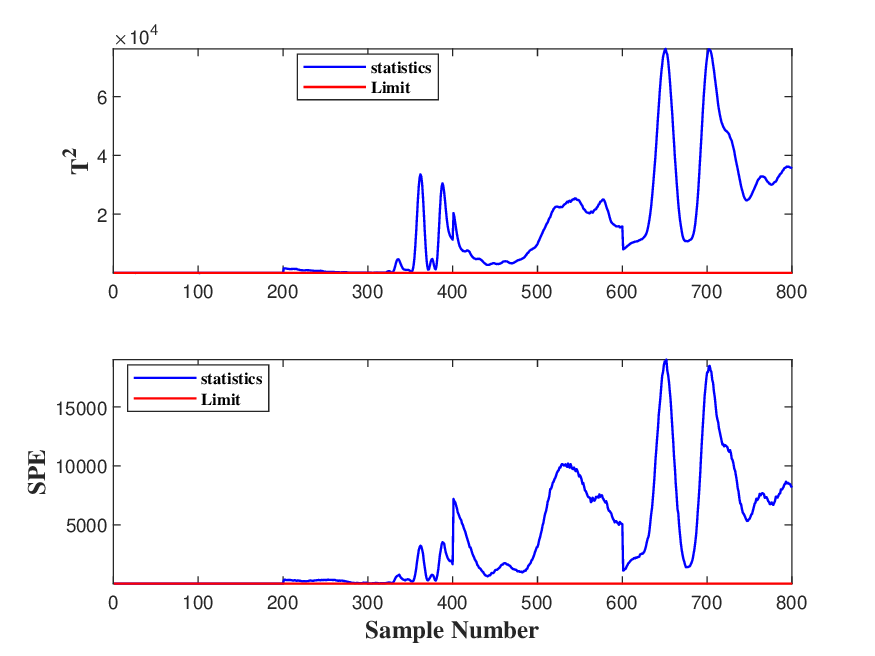}
	}
	\quad
	\subfigure[DiPCA.]{
		\includegraphics[width=0.45\textwidth,height=0.3\textwidth]{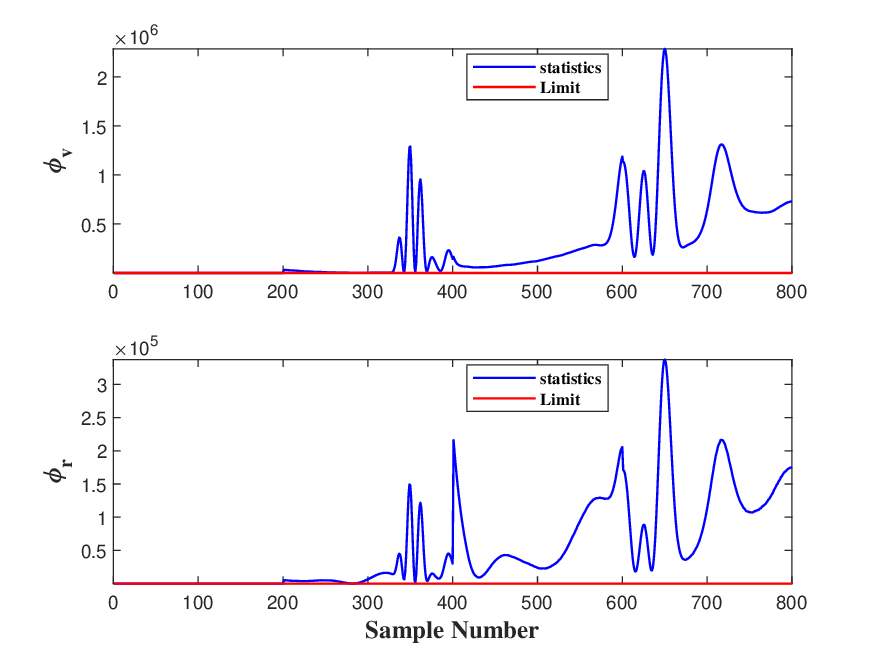}
	}
	\vspace{-1\baselineskip}
	\subfigure[DiCCA.]{
		\includegraphics[width=0.45\textwidth,height=0.3\textwidth]{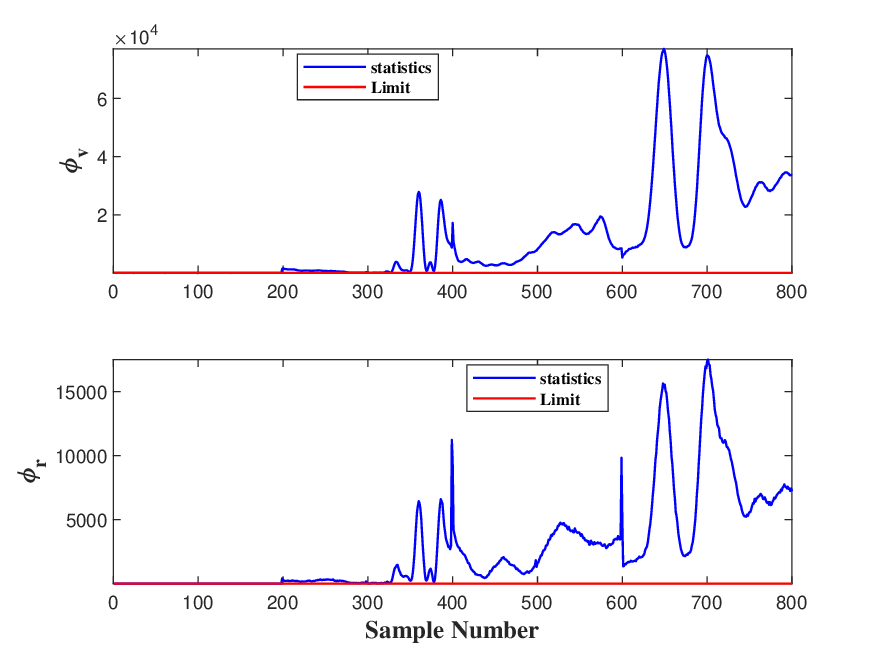}
	}
	\quad
	\subfigure[RDVDL.]{
		\includegraphics[width=0.45\textwidth,height=0.3\textwidth]{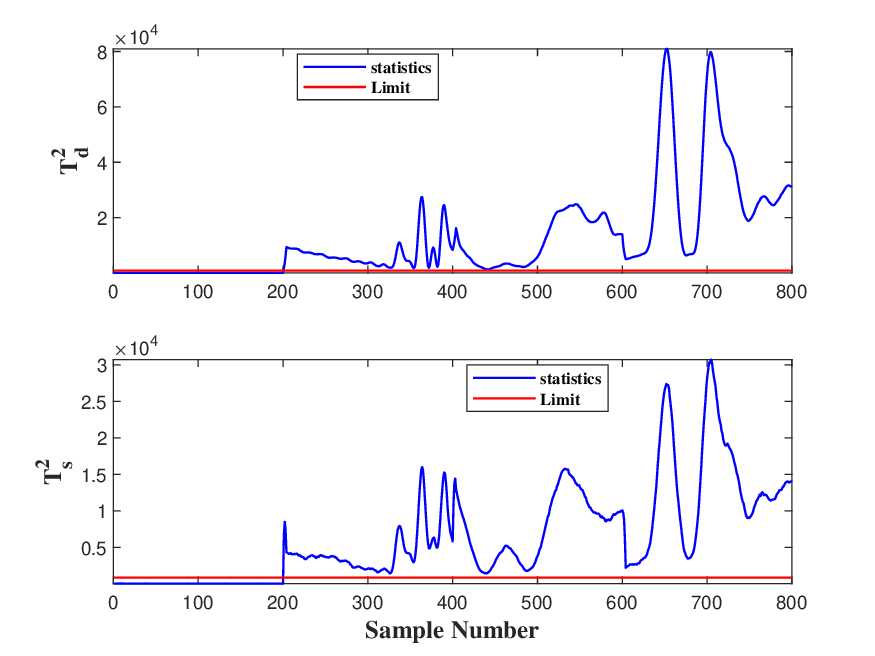}
	}
	\caption{Fault detection result of fault 2.}
	\label{fig.5}
\end{figure}
\begin{figure}[H]
	\centering
	\includegraphics[width=0.85\textwidth,height=0.5\textwidth]{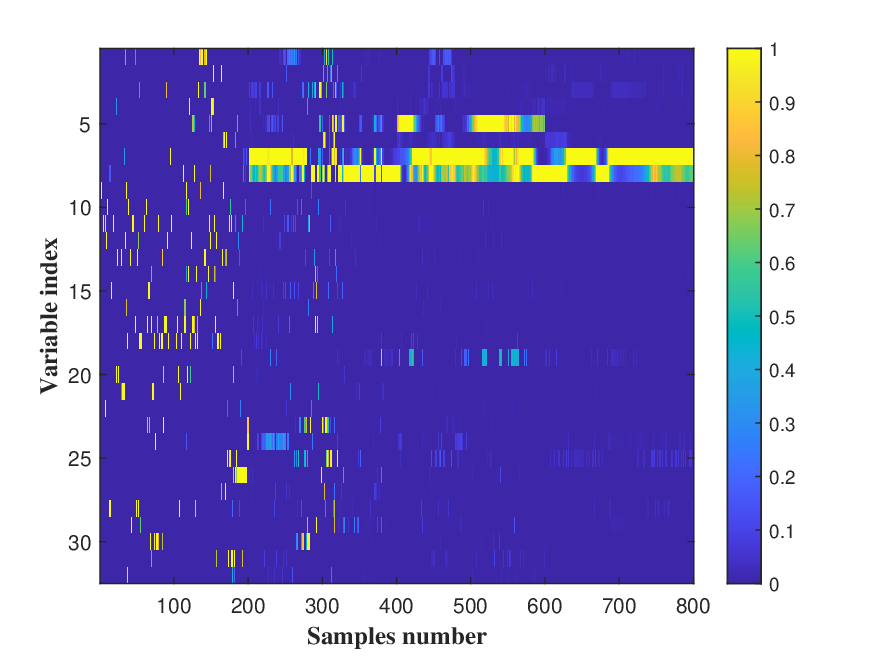}
	\caption{Fault diagnosis result of fault 2.}
	\label{fig.6}
\end{figure}

All four data-driven methods are also used to detect this fault, and the detection results are shown in Figure \ref{fig.5}. All four methods can detect the fault. The fault started from the 201st sample, but the change in the beginning of the fault was so small that it was not easy to detect. The RDVDL method established in this work has better sensitivity and can detect small fault changes. The fault results diagnosed by the RDVDL method are shown in Figure \ref{fig.6}. It can be seen that the variables 7 and 8 are the key factors for this fault.

\subsection{Increased electrolyzer temperature}

The electrolysis cell of the electrolyzer uses a diaphragm to isolate the hydrogen produced by the cathode and the oxygen produced by the anode, ensuring the purity of the electrolysis gas and the safety of the hydrogen production unit. 
Water electrolysis is an exothermic reaction and the higher the electrolysis current, the higher the temperature of the diaphragm. High temperatures can lead to changes in diaphragm chemistry, affecting hydrogen production performance and equipment safety.
The temperature of the electrolyzer is a core operating parameter that needs to be controlled in a hydrogen production system, and there can be several reasons for the increase in temperature. 
The main source of heat in the electrolyzer is the electrolysis process. As the current increases, the temperature rises. In addition, the circulation module is used to cool the electrolyzer, and abnormal operation of the circulation system can also cause the temperature of the electrolyzer to rise

\begin{figure}[H]
	\centering
	\subfigure[DPCA.]{
		\includegraphics[width=0.45\textwidth,height=0.3\textwidth]{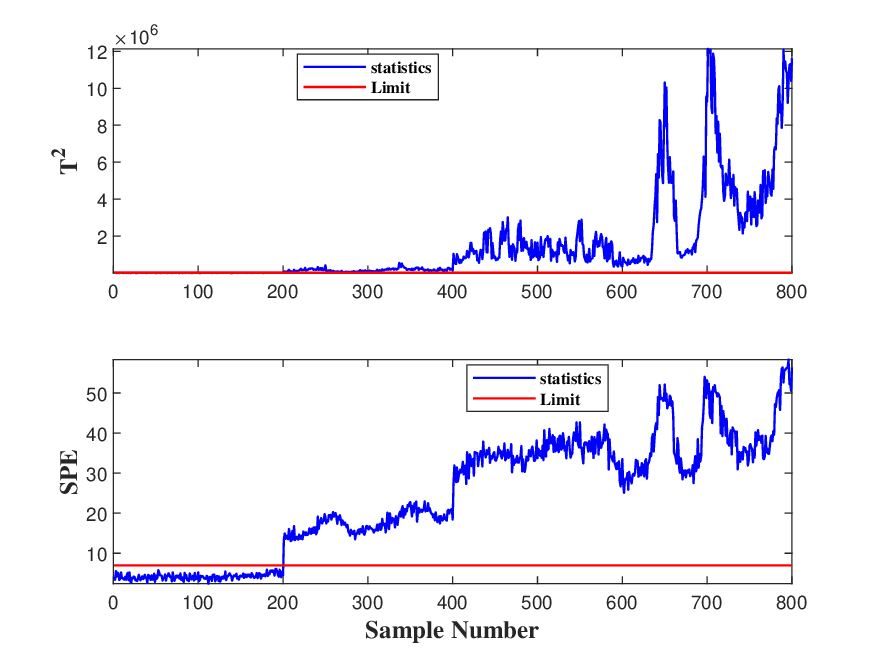}
	}
	\quad
	\subfigure[DiPCA.]{
		\includegraphics[width=0.45\textwidth,height=0.3\textwidth]{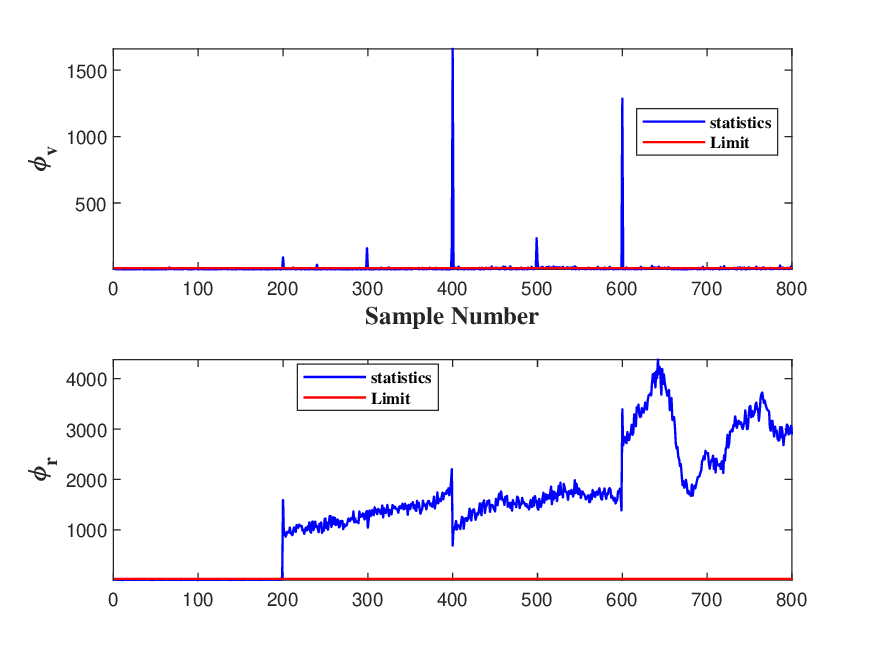}
	}
	\vspace{-1\baselineskip}
	\subfigure[DiCCA.]{
		\includegraphics[width=0.45\textwidth,height=0.3\textwidth]{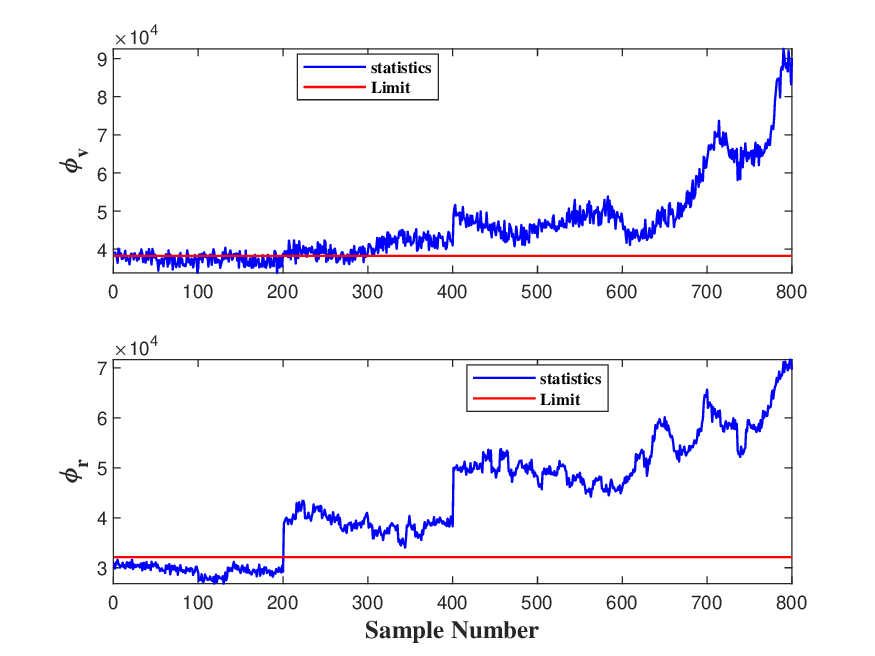}
	}
	\quad
	\subfigure[RDVDL.]{
		\includegraphics[width=0.45\textwidth,height=0.3\textwidth]{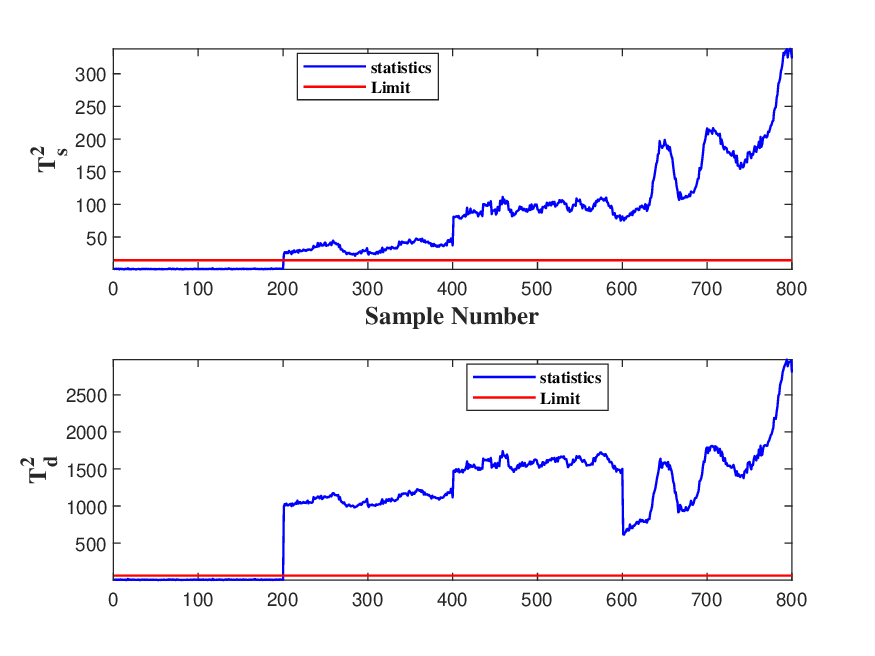}
	}
	\caption{Test result of fault 3.}
	\label{fig.7}
\end{figure}

\begin{figure}[H]
	\centering
	\includegraphics[width=0.85\textwidth,height=0.5\textwidth]{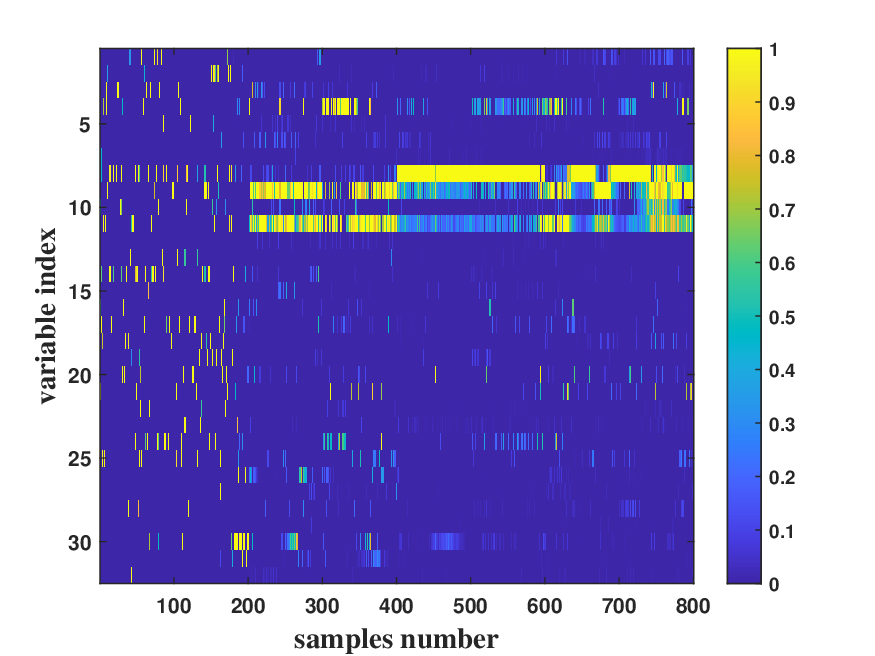}
	\caption{Monitoring result of fault 3.}
	\label{fig.8}
\end{figure}

Changes in these variables can be monitored using a data-driven method. The fault detection results for all four methods are shown in Figure \ref{fig.7}, where faults change more dramatically with time. Except for the DiPCA method, which did not detect changes in faults at the $ \phi_v $ statistic, all other methods detected a trend in fault changes. The RDVDL method proposed in this work can clearly detect changes in faults. The results of the diagnosis of fault 3 using the RDVDL method are shown in Figure \ref{fig.8}. Variables 8, 9 and 11 are diagnosed as the main cause of the fault.

\section{Conclusions} \label{sec:conclusions}

As low carbon becomes the new goal for industrial development, AWE hydrogen production has received higher attention. The scale of industrial hydrogen production is expanding, and safety is getting more and more attention and becoming a hot spot for research. With the development of sensor technology, there are many sensors for data collection in industrial hydrogen production process, and data-driven process monitoring technology has gained wide application. To date, there are few methods for data-driven modeling of AWE processes. In this study, we review the faults that exist in the AWE process and the variables that can be collected through different sensors. In addition, the application of a data-driven approach for dynamic modeling in the AWE hydrogen production process is briefly reviewed. This work develops a dynamic fault detection and diagnosis method for variable fractional Bayesian dictionary learning, and test results confirm that this method can not only detect faults in the AWE operation process, but also diagnose the variables causing the faults.

\section*{CRediT authorship contribution statement}

Qi Zhang: Investigation, Conceptualization, Data curation,
Methodology, Software, Writing – original draft. Shan Lu:
Supervision. Lei Xie: Conceptualization, Supervision, Writing – reviewing and
editing, Funding acquisition. Weihua Xu: Supervision. Hongye Su: Supervision, Funding acquisition.

\section*{Acknowledgements}

This work was supported in part by National Key R\&D Program of China (2018YFB1701102) and in part by National Natural Science Foundation of P.R. China (62073286) and inpart by Science Fund for Creative Research Groups of the National Natural Science Foundation of China (61621002).

\section*{Appendices}

\subsection*{VB-E step}

Due to the properties of the Bernoulli distribution, $ {{z}_{ik}} $ takes 1 and 0 with probability $ p $ and $ 1-p $ respectively.
\begin{align} \label{8}
	\begin{split}
		q({{z}_{ik}}=1)&\propto \exp [{{\mathbb{E}}_{q}}[\ln {{\pi }_{k}}]) \times \exp (-\frac{{{\gamma }_{\epsilon }}}{2}
		\{{{\mathbb{E}}_{q}}[s_{ik}^{2}]{{\mathbb{E}}_{q}}[\boldsymbol{d}_{k}^{\top}{\boldsymbol{d}_{k}}]\\
		&-2\boldsymbol{\mu} _{k}^{\top}{{\nu }_{ik}}{{\mathbb{E}}_{q}}[\boldsymbol{x}_{i}^{-k}]\}) \\ 
		q({{z}_{ik}}=0)&\propto \exp ({{\mathbb{E}}_{q}}[1-\ln {{\pi }_{k}}])  
	\end{split}
\end{align}

The expected value of $ {{z}_{ik}} $ can be calculated as
\begin{align} \label{9}
	{{\mathbb{E}}_{q}}[{{z}_{ik}}]={{\eta }_{ik}}=\frac{q({{z}_{ik}}=1)}{q({{z}_{ik}}=1)+q({{z}_{ik}}=0)}
\end{align}
where $ {{z}_{ik}}=1 $ if $ {{\eta }_{ik}} \ge 0.5  $ else $ {{z}_{ik}}=0 $.

In eq. \eqref{8}, the expectations of $ \ln ({{\pi }_{k}}) $ and $ \ln (1-{{\pi }_{k}}) $ can be calculated as
\begin{align}
	\begin{split}
		&{\mathbb{E}_{q}}[\ln ({{\pi }_{k}})]=\psi (\frac{{{a}_{0}}}{K}+\langle {{\eta}_{k}}\rangle )
		-\psi (\frac{{{a}_{0}}+{{b}_{0}}(K-1)}{K}+N) \\ 
		&{\mathbb{E}_{q}}[\ln (1-{{\pi }_{k}})]=\psi (\frac{{{b}_{0}}(K-1)}{K}+N
		-\langle {{\eta}_{k}}\rangle )\   
		-\psi (\frac{{{a}_{0}}+{{b}_{0}}(K-1)}{K}+N) \\ 
	\end{split}
\end{align}
where $ \psi(\cdot) $ represents the digamma function.
In Eq. \eqref{8}, $ {\mathbb{E}_{q}}[s_{ik}^{2}] $ and $ {\mathbb{E}_{q}}[\boldsymbol{d}_{k}^{\top}{\boldsymbol{d}_{k}}] $ can be calculated as
\begin{align}
	\begin{split}
		& {\mathbb{E}_{q}}[s_{ik}^{2}]={\langle\nu _{ik}\rangle}^{2}+{{\Omega }_{i}^{k}} \\ 
		& {\mathbb{E}_{q}}[\boldsymbol{d}_{k}^{\top}{\boldsymbol{d}_{k}}]={\langle\boldsymbol{\mu} _{k}^{\top}{\boldsymbol{\mu }_{k}}\rangle}+\text{trace}({\boldsymbol{\Sigma }_{k}}) \\
	\end{split} 
\end{align}
where $ \langle \eta_{k}\rangle=\Sigma_{i=1}^{N}\eta_{ik} $ is defined in the update for $ \boldsymbol{\pi} $, $ {{\Omega }_{i}^{k}} $ is the
$ k $-th diagonal element of  defined in the update for $ \boldsymbol{s} $.

\subsection*{VB-M step}

In the VB-M step, parameters are constantly updated.
The approximate posterior $ q(\boldsymbol{d}_{k}) $ can be obtained as
\begin{align} \label{12}
	\begin{split}
		& {{\boldsymbol{\mu }}_{k}}={{\gamma }_{\epsilon }}{{\mathbf{\Sigma }}_{k}}\underset{i=1}{\mathop{\overset{N}{\mathop{\sum }}\,}}\,{{\mathbb{E}}_{q}}[s_{ik}^{2}]{{\mathbb{E}}_{q}}[{{z}_{ik}}]{{\mathbb{E}}_{q}}[\boldsymbol{x}_{i}^{-k}] \\ 
		& {{\mathbf{\Sigma }}_{k}}={{\left( P{{\mathbf{I}}_{P}}+{{\gamma }_{\epsilon }}\underset{i=1}{\mathop{\overset{N}{\mathop{\sum }}\,}}\,{{\mathbb{E}}_{q}}[{{s}_{ik}}]{{\mathbb{E}}_{q}}[{{z}_{ik}}] \right)}^{-1}} 
	\end{split}
\end{align}

The approximate posterior $ q(s_{ik}) $ can be obtained as
\begin{align} \label{13}
	\begin{split}
		& {{\Omega}_{ik}}={{({{\gamma }_{s}}+{{\gamma }_{\epsilon }}{{\mathbb{E}}_{q}}[\boldsymbol{d}_{k}^{\top}{\boldsymbol{d}_{k}}]{{\mathbb{E}}_{q}}[{{z}_{ik}}])}^{-1}} \\ 
		& {{\nu }_{ik}}={{\gamma }_{\epsilon }}{{\Omega}_{ik}}\boldsymbol{\mu} _{k}^{\top}{{\mathbb{E}}_{q}}[{{z}_{ik}}]{{\mathbb{E}}_{q}}[\boldsymbol{x}_{i}^{-k}] 
	\end{split}
\end{align}

The approximate posterior $ q(\boldsymbol{\pi}) $ can be obtained as
\begin{align}\label{14}
	\begin{split}
		&{{\tau }_{1k}}=\frac{{{a}_{0}}}{K}+\left\langle {{n}_{k}} \right\rangle \\
		&{{\tau }_{2k}}=\frac{{{b}_{0}}(K-1)}{K}+N-\left\langle {{n}_{k}} \right\rangle
	\end{split}
\end{align}

The approximate posterior $ q(\gamma_{s}) $ and $ q(\gamma_{\epsilon}) $ have a form of a product of Gamma distributions, which can be obtained as follow
\begin{align}\label{15}
	\begin{split}
		& {c}'={{c}_{0}}+\frac{NK}{2}\quad {d}'={{d}_{0}}+\frac{1}{2}\underset{i=1}{\overset{N}{\mathop \sum }}\,\nu _{i}^{\top}{{\nu }_{i}} \quad {e}'={{e}_{0}}+\frac{NP}{2} \\  &{f}'={{f}_{0}}+\frac{1}{2}\underset{i=1}{\overset{{N}}{\mathop \sum }}\,\{{\boldsymbol{x}_{i}}\boldsymbol{x}_{i}^{\top}
		-2\underset{k=1}{\overset{K}{\mathop \sum }}\,\{{{\mathbb{E}}_{q}}[{{z}_{ik}}]{{\nu}_{ik}}\boldsymbol{\mu}_{k}^{\top}{\boldsymbol{x}_{i}}  
		+{{\mathbb{E}}_{q}}[({\boldsymbol{s}_{i}}\odot{\boldsymbol{z}_{i}}){{\boldsymbol{D}}^{\top}}\boldsymbol{D}{{({\boldsymbol{s}_{i}}\odot{\boldsymbol{z}_{i}})}^{\top}}]\}
	\end{split}  
\end{align}

where
\begin{align}\label{16}
	\begin{split}
		& {{\mathbb{E}}_{q}}[({\boldsymbol{s}_{i}}\odot{\boldsymbol{z}_{i}}){{\boldsymbol{D}}^{\top}}\boldsymbol{D}{{({\boldsymbol{s}_{i}}\odot{\boldsymbol{z}_{i}})}^{\top}}]\\
		&=\sum\limits_{k=1}^{K}{\sum\limits_{k\ne {{k}^{\prime }}}}
		{{\nu }_{ik}}{{\mathbb{E}}_{q}}[{{z}_{ik}}]{{\nu }_{ik}^{\prime }}{{\mathbb{E}}_{q}}[{{z}_{ik}}]^{\prime }
		\boldsymbol{\mu} _{k}^{\top}{\boldsymbol{\mu }_{{{k}^{\prime }}}} 
		+\sum\limits_{k=1}^{K}{{\mathbb{E}}_{q}}[{{z}_{ik}}]{{E}_{q}}[s_{ik}^{2}]{{E}_{q}}[\boldsymbol{d}_{k}^{\top}{\boldsymbol{d}_{{{k}^{\prime }}}}] 
	\end{split} 
\end{align}


\bibliography{reference}

\bibliographystyle{elsarticle-num}

\end{document}